\documentclass[letterpaper, 10 pt, conference]{ieeeconf} 

\IEEEoverridecommandlockouts                              
\usepackage{times}

\usepackage{cite}
\usepackage{multicol}
\usepackage[bookmarks=true]{hyperref}
\usepackage{amsmath}
\usepackage[table]{xcolor} 
\usepackage{colortbl}
\usepackage{graphicx}
\usepackage{caption}
\usepackage{cleveref}
\usepackage{bm}
\usepackage{gensymb}
\usepackage{booktabs}
\usepackage{diagbox}
\usepackage{multirow}
\usepackage[table]{xcolor}
\usepackage{pgf}
\usepackage{float}
\definecolor{ourgray}{gray}{0.9}
\usepackage[table]{xcolor}
\definecolor{projectblue}{RGB}{0,82,155}
\definecolor{normgreen}{RGB}{0,160,0}
\definecolor{0p0}{RGB}{0, 80, 0}
\definecolor{3p9}{RGB}{0, 85, 0}
\definecolor{8p3}{RGB}{0, 92, 0}
\definecolor{10p0}{RGB}{0, 94, 0}
\definecolor{20p0}{RGB}{0, 108, 0}
\definecolor{25p0}{RGB}{0, 115, 0}
\definecolor{30p0}{RGB}{0, 122, 0}
\definecolor{35p0}{RGB}{0, 129, 0}
\definecolor{35p7}{RGB}{0, 130, 0}
\definecolor{36p5}{RGB}{0, 131, 0}
\definecolor{40p0}{RGB}{0, 136, 0}
\definecolor{41p4}{RGB}{0, 138, 0}
\definecolor{45p0}{RGB}{0, 143, 0}
\definecolor{47p5}{RGB}{0, 147, 0}
\definecolor{48p0}{RGB}{0, 147, 0}
\definecolor{50p0}{RGB}{0, 150, 0}
\definecolor{55p0}{RGB}{0, 157, 0}
\definecolor{59p8}{RGB}{0, 164, 0}
\definecolor{60p5}{RGB}{0, 165, 0}
\definecolor{65p0}{RGB}{0, 171, 0}
\definecolor{70p0}{RGB}{0, 178, 0}
\definecolor{73p7}{RGB}{0, 183, 0}
\definecolor{75p0}{RGB}{0, 185, 0}
\definecolor{80p0}{RGB}{0, 192, 0}
\definecolor{85p0}{RGB}{0, 199, 0}
\definecolor{90p0}{RGB}{0, 206, 0}
\definecolor{94p0}{RGB}{0, 212, 0}
\definecolor{96p0}{RGB}{0, 214, 0}
\definecolor{100}{RGB}{0, 220, 0}  
\definecolor{c1}{RGB}{130,20,20}    
\definecolor{c2}{RGB}{155,40,28}    
\definecolor{c3}{RGB}{178,62,32}    
\definecolor{c4}{RGB}{196,88,36}    
\definecolor{c5}{RGB}{204,116,40}   
\definecolor{c6}{RGB}{200,144,44}   
\definecolor{c7}{RGB}{184,166,50}   
\definecolor{c8}{RGB}{158,176,58}   
\definecolor{c9}{RGB}{120,165,68}   
\definecolor{c10}{RGB}{70,140,60}   
\definecolor{c11}{RGB}{8,105,42}    

\definecolor{g9}{RGB}{238, 192, 181}
\definecolor{g8}{RGB}{246, 206, 190}
\definecolor{g7}{RGB}{252, 220, 198}
\definecolor{g6}{RGB}{254, 232, 204}
\definecolor{g5}{RGB}{254, 236, 205}
\definecolor{g4}{RGB}{255, 250, 230} 
\definecolor{g3}{RGB}{235, 240, 200} 
\definecolor{g2}{RGB}{200, 225, 170} 
\definecolor{g1}{RGB}{155, 205, 135} 

\newcommand{\monoicon}{\raisebox{-0.15\height}{\includegraphics[height=.9em]{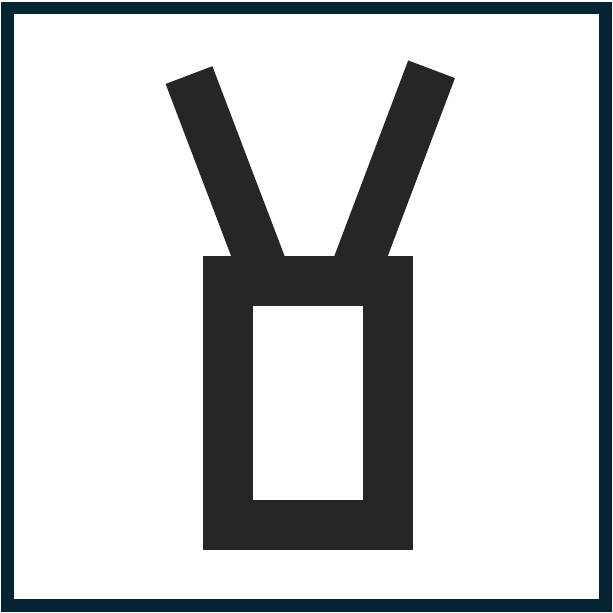}}}
\newcommand{\stereoicon}{\raisebox{-0.15\height}{\includegraphics[height=.9em]{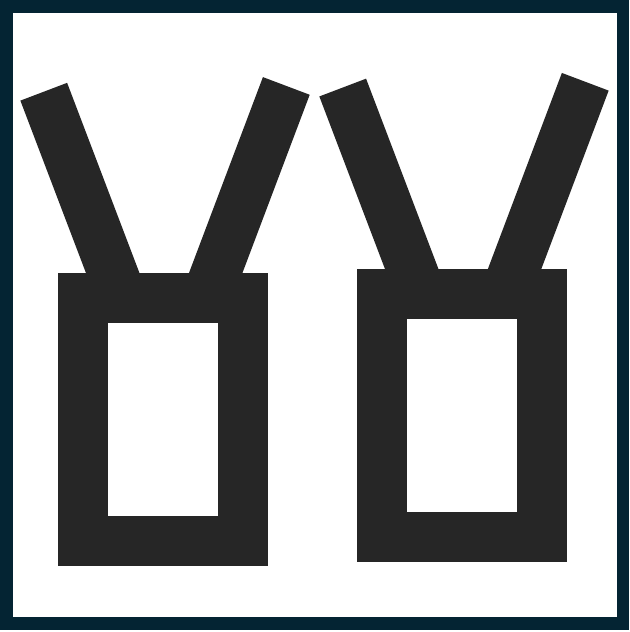}}}
\newcommand{\argmin}{\operatornamewithlimits{argmin}}

\definecolor{srgray}{gray}{0.6} 
\newcommand{\res}[2]{%
  $\displaystyle #1$\kern0.1em%
  {\textcolor{black!55}{\scalebox{0.85}{(#2)}}}%
}

\usepackage{amssymb,amsfonts}
\usepackage[caption=false]{subfig}

\DeclareMathOperator{\Exp}{Exp}
\DeclareMathOperator{\Log}{Log}
\newcommand{\SO}{\mathrm{SO}}
\newcommand{\SE}{\mathrm{SE}}
\newcommand{\R}{\mathbb{R}}
\newcommand{\bI}{\mathbf{I}}
\newcommand{\bzero}{\mathbf{0}}
\newcommand{\what}[1]{#1^{\wedge}}
\newcommand{\Ad}{\mathrm{Ad}}

\begin{document}

\title{\LARGE \bf
MAC-I$^2$: Learned Metrics-Aware Covariance for Robust Visual-Inertial Fusion in Initialization and Calibration \\
{\small\normalfont\sffamily\bfseries
\href{https://mac-i2.github.io/}{
  \textcolor{projectblue}{Project Page: mac-i2.github.io}
}}
\vspace{-0.6em}
}

\author{
Xiang Fei$^{1,*}$,
Yuheng Qiu$^{1,*,\ddagger}$,
Can Xu$^{1,3}$,
Yutian Chen$^{1}$,
Ruogu Li$^{1}$,
\\
Xingxing Zuo$^{2}$,
Wenshan Wang$^{1}$,
Sebastian Scherer$^{1}$
\thanks{$^{*}$ Equal contribution.}
\thanks{$^{\ddagger}$ Project lead.}
\thanks{$^{1}$ The Robotics Institute, Carnegie Mellon University,
5000 Forbes Ave, Pittsburgh, PA 15213 USA.}
\thanks{$^{2}$ Mohamed bin Zayed University of Artificial Intelligence
(MBZUAI), Masdar City, Abu Dhabi, UAE.}
\thanks{$^{3}$ University of Toronto Institute for Aerospace Studies (UTIAS),
University of Toronto, 4925 Dufferin St, Toronto, ON M3H 5T6, Canada.}
}




%

\thispagestyle{empty}
\pagestyle{empty}
\makeatletter
\g@addto@macro\@maketitle{%
  \par\vspace{0.8em}
  \captionsetup{type=figure}\setcounter{figure}{0}
  \def\mycolspace{1.2mm}
  \centering
\includegraphics[width=1.78\columnwidth]{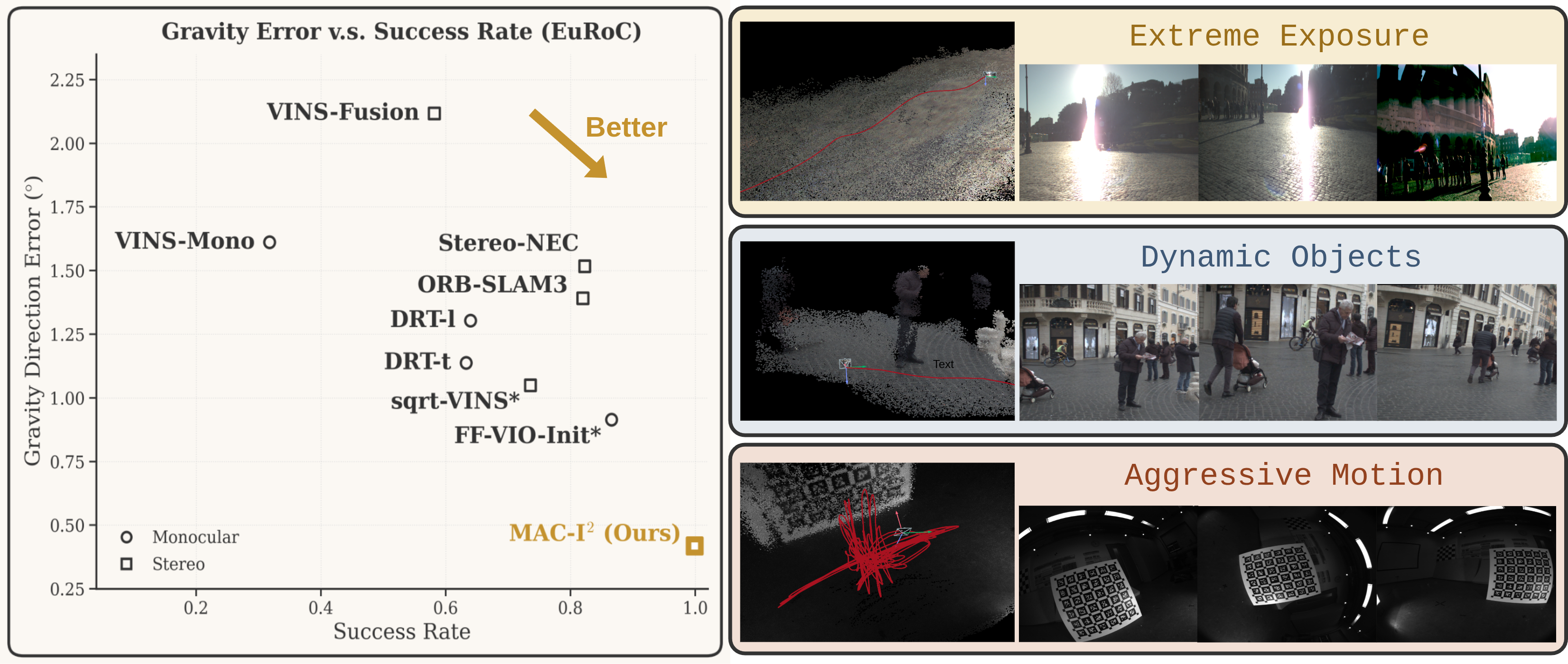}
	\captionof{figure}{MAC-I$^2$ achieves robust VI fusion through learned metrics-aware covariances for both visual and inertial measurements. We build a VI initialization and calibration system, which substantially outperforms existing methods (left).
    } 
	\label{fig:figure1}
  \vspace{-5pt}
}
\makeatother

\maketitle

\begin{abstract} 
Visual-Inertial (VI) fusion is fundamental to accurate and robust state estimation, where camera and IMU measurements are combined according to their respective uncertainties. Existing methods, however, fuse the two modalities with predefined uncertainties, regardless of how reliable each is in the local context, and thus often struggle under challenging environments involving illumination changes, dynamic objects, and textureless regions. In this paper, we present MAC-I$^2$, which achieves robust VI fusion through learned metric-aware covariance for both modalities, so that vision and IMU compete on their own merits rather than relying on predefined uncertainties. Here, metrics-aware means that each predicted covariance faithfully reflects the actual magnitude of the corresponding measurement noise. On the visual side, we propagate learned feature-matching uncertainties into pose covariances for the fusion. On the inertial side, motivated by the observation that integration error accumulates sharply at the early stage and grows slowly afterward, we design a learned IMU model with a learnable initial covariance, and propose a dedicated fine-tuning strategy on a held-out training subset to enable the metrics-aware covariance on unseen sequences. As a showcase, we build a VI initialization and calibration system, since accurate and robust initialization and calibration are the prerequisite for any reliable VI system. Experiments on EuRoC, and VBR show that MAC-I$^2$ substantially outperforms existing methods: it achieves a 99.9\% initialization success rate on EuRoC, reducing gravity and velocity errors by about 60\% and 42\% over the strongest baseline, and maintains 80\% success rate on challenging VBR sequences where baseline methods such as VINS-Mono drop below 10\%.

\end{abstract}

\IEEEpeerreviewmaketitle

\section{Introduction}
\label{sec:introduction}

Visual-inertial (VI) fusion is a cornerstone technology for autonomous systems, AR/VR, and robotics, providing accurate and robust state estimation by combining complementary visual and inertial measurements~\cite{qin2018vins, campos2021orb, leutenegger2015keyframe}. At its core, VI fusion weights the two modalities according to how much each should be trusted, where the trust is encoded by the covariance assigned to each measurement. In practice, the reliability of each modality varies with the local context: visual measurements degrade under illumination changes, dynamic objects, and textureless regions, while inertial reliability is affected by the motion patterns. However, existing methods rely on predefined covariances that fail to capture the reliability of each measurement.

This reliance on predefined covariances is evident across existing VI systems. Factor-graph-based estimators such as VINS-Mono~\cite{qin2018vins} and OKVIS~\cite{leutenegger2015keyframe} weight visual residuals by a fixed pixel noise (e.g. 1.5 pixels in VINS-Mono), while inertial residuals are weighted by covariances propagated from preset IMU noise parameters. ORB-SLAM3~\cite{campos2021orb} assigns the visual information matrix according to the image pyramid level at which each feature is detected, so that features from the same scale share an identical covariance. Filter-based estimators such as MSCKF~\cite{msckf} and OpenVINS~\cite{geneva2020openvins} likewise rely on a fixed measurement-noise covariance in the EKF update. In all these cases, the covariance follows a predefined rule rather than the actual reliability of each measurement in context, and therefore cannot reflect the varying conditions encountered at deployment. As a result, unreliable measurements might be over-weighted during fusion, leading to degraded accuracy or even complete failure.

Recent progress in learning-based methods~\cite{qiu2025mac, qiu2023airimu} offers a promising alternative by learning uncertainty directly from data. MAC-VO~\cite{qiu2025mac} learns \emph{metrics-aware} feature-matching uncertainty, where each predicted uncertainty faithfully reflects the actual magnitude of the matching noise and thus quantitatively indicates how reliable each match is. AirIMU~\cite{qiu2023airimu} learns the per-frame uncertainty of inertial measurements and propagates it into the uncertainty of IMU integration, replacing fixed noise parameters with data-driven estimates. These advances suggest that learned, metrics-aware uncertainties could enable fusion that weights each measurement by its reliability in local contexts. However, the inertial uncertainty from AirIMU is not metrics-aware, as we show later, and more importantly, no work has yet exploited learned metrics-aware covariances for VI fusion.

In this paper, we propose MAC-I$^2$, which achieves robust VI fusion through learned metrics-aware covariances for both visual and inertial measurements, so that vision and IMU compete on their own merits rather than relying on predefined uncertainties. For the visual modality, we propagate the learned feature-matching uncertainties from MAC-VO into pose covariances through the information matrix at convergence of the visual pose estimation, then bringing metrics-aware visual uncertainty into the fusion. For the inertial modality, we observe that the integration error accumulates sharply at the early stage and grows slowly afterward, and accordingly design a learned IMU model with a learnable initial covariance to capture this pattern. In addition, we further introduce a fine-tuning strategy on a held-out training subset, enabling metrics-aware inertial covariances on unseen sequences. As a showcase, we apply MAC-I$^2$ to VI initialization and calibration, which are the prerequisite for running VI pose estimation. The main contributions are:

\begin{itemize} 
    \item We propose MAC-I$^2$, the first framework to fuse vision and IMU through learned metrics-aware covariances, weighting each measurement by its reliability in local contexts instead of predefined uncertainties.

    \item We derive metrics-aware visual pose covariances from learned feature-matching uncertainties, and design a learned IMU model with a learnable initial covariance together with a held-out fine-tuning strategy that yields metrics-aware inertial covariances on unseen sequences. Ablation studies and analysis validate their effectiveness and metrics-aware properties.

    \item As a showcase, we build a VI initialization and calibration system, demonstrating substantial improvement over existing methods, including a 99.9\% initialization success rate on EuRoC benchmark and a 80\% success rate on challenging VBR sequences where baseline methods such as VINS-Mono drop below 10\%.

\end{itemize}


\section{Related Work}
\label{sec:related_work}

\subsection{Traditional Uncertainty Modeling in VI Fusion}

Covariances are central to VI fusion, as they determine how much each visual and inertial measurement is trusted, and a range of strategies have been explored to set them. The most common choice is to assign fixed or hand-tuned covariances to visual and inertial residuals~\cite{qin2018vins, leutenegger2015keyframe, campos2021orb, geneva2020openvins, zhang2026efficient}, which are simple but cannot reflect how reliable each measurement is in local contexts. To introduce adaptivity, robust cost functions and back-ends such as switchable constraints~\cite{switchable} and dynamic covariance scaling~\cite{agarwal2013dcs} down-weight residuals with large errors during optimization, while heuristic schemes adapt the visual covariance according to indicators such as image quality or feature counts~\cite{asil2025}. On the inertial side, adaptive filters online-estimate time-varying IMU noise covariances~\cite{akflio}, and online IMU self-calibration estimates noise and intrinsic parameters during operation~\cite{xiao_imu_selfcalib}. While these strategies relax the rigidity of fixed covariances, the resulting weights are still governed by heuristics or residual statistics rather than the measurement uncertainty, and therefore still fail to capture how reliable each measurement actually is.

\subsection{VI Initialization and Calibration}

VI initialization and calibration provide the initial states and camera-IMU extrinsics required for VI pose estimation. Existing methods vary in how they allocate trust between vision and IMU. A common strategy first estimates camera motion from visual correspondences and then fully trusts these poses to align IMU integration~\cite{qin2018vins, mur2017visual}, which is efficient but fragile, as erroneous visual poses directly corrupt the alignment. To reduce this dependence, decoupled methods trust bias-corrected gyroscope integration for rotation and vision for translation~\cite{he2023rotation}, while others jointly optimize visual and inertial residuals~\cite{campos2021orb, mu2025robust}, where the relative weighting between them is still set by predefined covariances. For extrinsic calibration, traditional methods rely on offline procedures with structured targets~\cite{kalibr, yang2024multi}, and online approaches integrate calibration into the estimator for target-free operation~\cite{huang2018online, yang2023online}, but their reliance on geometric correspondences and hand-tuned noise models makes calibration brittle. In all cases, the trust between vision and IMU is predetermined rather than adapted to how reliable each measurement actually is, which motivates our use of learned metrics-aware covariances to let the two modalities compete on their own merits.

\subsection{Learning-based Uncertainty Modeling}
Learning-based methods offer an alternative by learning sensor uncertainty directly from data. For vision, learned uncertainty has been used to weight residuals in direct visual odometry~\cite{yang2020d3vo}, while MAC-VO~\cite{qiu2025mac} predicts a \emph{metrics-aware} covariance for each feature, where the predicted uncertainty reflects the actual magnitude of the matching noise rather than only the relative confidence among features. For the inertial modality, TLIO~\cite{tlio} is a learned inertial odometry, which regresses 3D displacement together with the corresponding uncertainty. 
VIO-DualProNet~\cite{SOLODAR2024108466} and AirIMU~\cite{qiu2023airimu} instead learn the uncertainty of the inertial measurements and propagate it into the integration covariance, which can be used for fusion, yet this learned covariance is not metrics-aware. Consequently, to the best of our knowledge, we are the first to bring metrics-aware covariances from both vision and IMU into a common fusion. MAC-I$^2$ fills this gap by producing metrics-aware covariances for both modalities and fusing them in a joint optimization.

\section{Notations and Preliminaries}
\subsection{Notations}
As a showcase, we apply MAC-I$^2$ to VI initialization and calibration, for which we define the state:
\begin{align}
\mathcal {X}=&\big \{ \mathbf {x}_{0},\mathbf {x}_{1},\ldots \mathbf {x}_{n}, \mathbf{g}^{c_{0}}, \mathbf{b}_g, \mathbf{b}_a, \mathbf {T}^{b}_{c} \big \} \notag, \\ \label{formula:state} \mathbf {x}_{k}=&\big \{ \mathbf {T}^{c_{0}}_{b_{k}},\mathbf {v}^{b_{k}}_{b_{k}} \big \},
\end{align}
where $\mathbf{T}^{c_0}_{b_k} \in \operatorname{SE}(3)$ denotes the pose of the $k$-th IMU frame with respect to the first camera frame $\{c_0\}$. $\mathbf{v}^{b_k}_{b_k}$ denotes the velocity of the $k$-th IMU frame expressed in its local frame. $\mathbf{g}^{c_0}$ denotes the gravity vector expressed in the first camera frame. $\mathbf{b}_g$ and $\mathbf{b}_a$ denote the IMU gyroscope and accelerometer biases, respectively, which are assumed to remain constant during the initialization and calibration process. $\mathbf{T}^b_c$ denotes the extrinsic from the camera frame to the IMU frame. For initialization with known camera-imu extrinsics, $\mathbf{{T}}^{b}_{c}$ can be removed from the state variables.

\subsection{IMU Integration and learning-based IMU model}
The IMU integration follows the standard approach as proposed in \cite{forster_imu}:
\begin{equation}
\label{formula:imu_integration}
\begin{aligned}
\mathbf{q}^{c_0}_{b_j} &= \mathbf{q}^{c_0}_{b_i} \bm{\gamma}_{b_j}^{b_i}, \\
\mathbf{v}_{b_j}^{c_0} &= \mathbf{v}_{b_i}^{c_0} + \mathbf{g}^{c_0} \Delta t_{ij} + \mathbf{R}^{c_0}_{b_i} \bm{\beta}_{b_j}^{b_i}, \\
\mathbf{p}^{c_0}_{b_{j}} &= \mathbf{p}^{c_0}_{b_i} + \mathbf{v}^{c_0}_{b_i} \Delta t_{ij} + \frac{1}{2} \mathbf{g}^{c_0} \Delta t_{ij}^2 + \mathbf{R}^{c_0}_{b_i} \bm{\alpha}_{b_j}^{b_i},
\end{aligned}
\end{equation}
where $\mathbf{q}_{b_i}^{c_0} \in \operatorname{SO}(3)$ denotes the rotation quaternion from the $i$-th body frame to the initial camera frame $c_0$, and $\mathbf{R}^{c_0}_{b_i}$ denotes its matrix form. $\bm{\gamma}_{b_j}^{b_i}$, $\bm{\beta}_{b_j}^{b_i}$, and $\bm{\alpha}_{b_j}^{b_i}$ denote IMU preintegration terms for rotation, velocity, and translation between time $i$ and $j$, which can be computed with the raw IMU measurements $\left\{\, \mathbf{a}_k,\ \bm{\omega}_k \ \middle|\ k = i, i+1, \dots, j \,\right\}$:
\begin{equation}
\begin{aligned}
\bm{\gamma}_{b_j}^{b_i} &= \prod\nolimits_{k=i}^{j-1} \text{Exp} \left({\bm{\omega}}_k \Delta t \right), \\
\bm{\beta}_{b_j}^{b_i} &= \sum\nolimits_{k=i}^{j-1} \mathbf{R}^{b_i}_{b_k} \mathbf{a}_k \Delta t,\\
\bm{\alpha}_{b_j}^{b_i} &= \sum\nolimits_{k=i}^{j-1} \left( \bm{\beta}_{b_k}^{b_i} \Delta t + \frac{1}{2} \mathbf{R}^{b_i}_{b_k} \mathbf{a}_k \Delta t^2 \right),
\end{aligned}
\end{equation}

AirIMU \cite{qiu2023airimu} is a learned model to estimate the corrections $[\bm{\sigma}^{\text{acc}},\bm{\sigma}^{\text{gyro}}]$ and the uncertainty $[\bm{\eta}^{\text{acc}}, \bm{\eta}^{\text{gyro}}]$ of the IMU measurements, with the input as the raw IMU data $[\mathbf{a}, \bm{\omega}]$:
\begin{equation}
\label{formula:correction_uncertainty}
\begin{aligned}
(\bm{\sigma}^{\text{acc}},\bm{\sigma}^{\text{gyro}}) = g_{\bm{\theta}} (\mathbf{a},\bm{\omega})&, \quad (\bm{\eta}^{\text{acc}}, \bm{\eta}^{\text{gyro}}) = h_{\bm{\theta}} (\mathbf{a},\bm{\omega}), \\
\tilde{\mathbf{a}} = \mathbf{a} + \bm{\sigma}^{\text{acc}}&, \quad \tilde{\bm{\omega}} = \bm{\omega} + \bm{\sigma}^{\text{gyro}},
\end{aligned}
\end{equation}
where $\bm{\theta}$ is the parameter of the neural network. The corrected measurements $\tilde{\mathbf{a}}$ and $\tilde{\bm{\omega}}$ are obtained by adding the learned corrections $\bm{\sigma}$ to the raw measurements, while the predicted uncertainties $\bm{\eta}$ characterize the measurement noise.

\subsection{Learning-based Visual Odometry MAC-VO}
MAC-VO \cite{qiu2025mac} is a learned visual odometry that learns a metrics-aware uncertainty model to select reliable keypoints and weight residuals in pose graph optimization (PGO).

Specifically, MAC-VO estimates camera poses by solving the following PGO for consecutive frames $c_k$ and $c_{k+1}$:
\begin{equation}
\begin{aligned}
    \mathbf{T}^{c_{k}}_{c_{k+1}}{}^\star &= \argmin_{\mathbf{T}^{c_{k}}_{c_{k+1}}}{\sum_{i}{\left\| \mathbf{p}_{k, i}^{c} - \mathbf{T}^{c_{k}}_{c_{k+1}} \mathbf{p}^{c}_{k+1, i} \right\|^2_{\bm{\Sigma}_i}}}, \\
    \bm{\Sigma}_i &= \bm{\Sigma}^p_{k,i} + \mathbf{R}^{c_{k}}_{c_{k+1}} \bm{\Sigma}^p_{k+1,i} (\mathbf{R}^{c_{k}}_{c_{k+1}})^T,
\end{aligned}
\label{formula:macvo}
\end{equation}
where $\mathbf{p}^{c}_{k, i}$ denotes the $i$-th keypoint in frame $c_k$, and $\|\cdot\|_{\bm{\Sigma}_i}$ represents the Mahalanobis distance with covariance matrix $\bm{\Sigma}_i$. $\bm{\Sigma}^p_{k,i}$ denotes the learned 3D covariance of $\mathbf{p}^{c}_{k, i}$.


\section{Methodology}
MAC-I$^2$ achieves robust VI fusion by predicting metrics-aware covariances for both modalities, so that visual and inertial measurements are weighted by their reliability in local contexts. We model metrics-aware covariance for the estimated visual poses (\cref{subsec:visual_cov}) and predict metrics-aware covariance for the IMU integration through a learned IMU model (\cref{subsec:imu_cov}). These two covariances are then combined in a VI initialization and calibration system (\cref{subsec:init_calib}) as a showcase application, where they jointly weight the visual and inertial residuals in the optimization.

\subsection{Metrics-Aware Visual Pose Covariance}
\label{subsec:visual_cov}

MAC-VO~\cite{qiu2025mac} provides a learned, metrics-aware 3D uncertainty for each visual feature. 
Many VI tasks, however, do not fuse raw features but first estimate the visual pose and then fuse it with the IMU, as in VI initialization, where directly optimizing raw measurements is unstable without good initial values. We therefore derive the covariance of estimated visual poses from the learned 3D feature uncertainty.
Notably, despite being trained solely on synthetic TartanAir \cite{wang2020tartanair} dataset, our covariance model produces metrics-aware pose uncertainties on real-world datasets without any fine-tuning.

The covariance of the estimated pose $\bm{\Sigma}_{\mathbf{T}^{c_{k}}_{c_{k+1}}}$ can be approximated via the Jacobian computed at convergence of the PGO problem in \eqref{formula:macvo}, where $k$ indexes the frames and $i$ indexes the keypoints matched between frames $c_{k}$ and $c_{k+1}$. We consider the left perturbation model $\mathbf{r}_{i,k}(\delta\bm{\xi})=\mathbf{p}_{k, i}^{c} - \text{Exp}(\delta\bm{\xi})\mathbf{T}^{c_{k}}_{c_{k+1}} \mathbf{p}^{c}_{k+1, i}$, which yields:
\begin{equation}
\begin{aligned}
    \mathbf{J}_{i,k} &= \frac{\partial{\mathbf{r}_{i,k}(\delta\bm{\xi})}}{\partial \delta\bm{\xi}} = \begin{bmatrix} -\mathbf{I}_{3\times3} & (\mathbf{T}^{c_{k}}_{c_{k+1}}\mathbf{p}^c_{k+1,i})^{\wedge} \end{bmatrix}, \\
    \mathbf{H}_{k} &\approx \sum_i (\mathbf{J}_{i,k})^T \bm{\Sigma}_i^{-1} \mathbf{J}_{i,k}, \quad
     \bm{\Sigma}_{\mathbf{T}^{c_{k}}_{c_{k+1}}} \approx \mathbf{H}_{k}^{-1},
\end{aligned}
\label{pose_cov}
\end{equation}
where $\mathbf{J}_{i,k}$ is the left Jacobian of the residual w.r.t. the left perturbation $\delta\bm{\xi}$ in the tangent space of the estimated visual pose $\mathbf{T}^{c_{k}}_{c_{k+1}}$, $(\cdot)^{\wedge}$ denotes the skew matrix of a vector, and $\bm{\Sigma}_i$ denotes the covariance matrix defined in \eqref{formula:macvo}. 

\begin{figure}[htbp]
\centering
\includegraphics[width=\columnwidth]{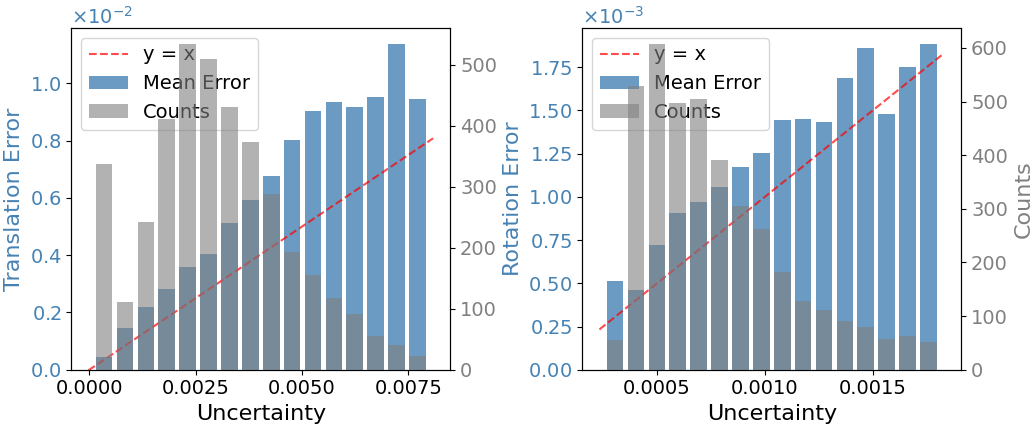}
\caption{Translation and rotation pose errors versus the predicted visual pose covariance on EuRoC. Blue bars denote the mean error within each bin, and grey bars denote the sample counts. The estimated uncertainty exhibits metrics-aware behavior, closely following the optimal $y = x$.}
\label{fig:pose_cov}
\vspace{-10pt}
\end{figure}

The estimated visual pose covariance is metrics-aware, as illustrated in Fig. \ref{fig:pose_cov}. 
Specifically, we group the pose estimations into bins according to their predicted uncertainty and compute the mean translation and rotation error within each bin, following the method in \cite{veicht2024geocalib}. 
The results show that the estimated uncertainty closely follows the optimal relationship of $y = x$, indicating a strong correlation between predicted uncertainty and actual pose error.

\subsection{Metrics-Aware Inertial Covariance}
\label{subsec:imu_cov}

\begin{figure}[tbp]
\centering
\includegraphics[width=\columnwidth]{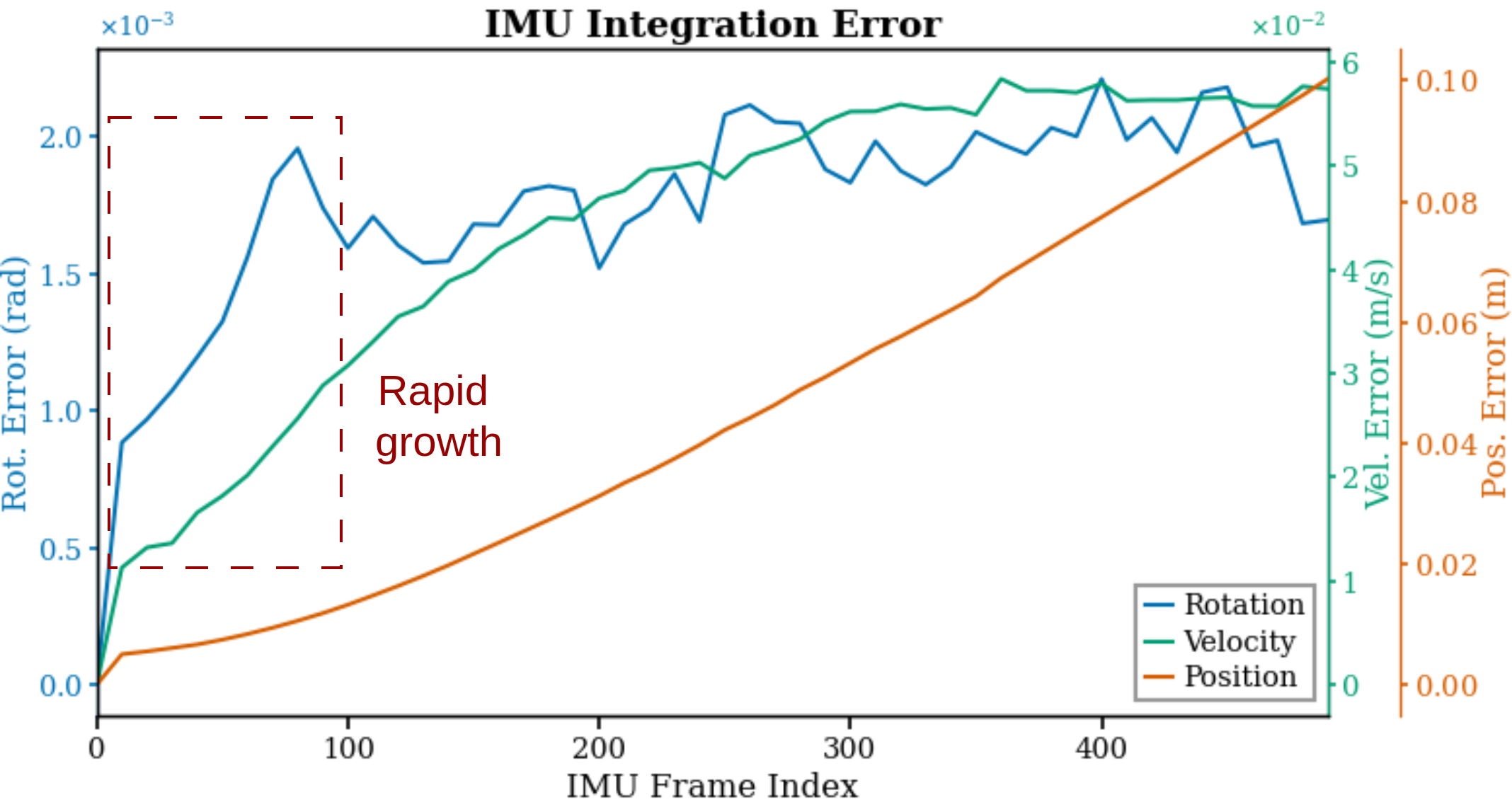}
\caption{Average IMU integration error with ground-truth bias correction on the EuRoC \texttt{MH\_04} sequence. 
}
\label{fig:inte_error}
\vspace{-5pt}
\end{figure}


\noindent \textbf{Observation: after bias correction, the IMU integration error rises sharply at the start of the window and accumulates slowly thereafter.}
As shown in Fig.~\ref{fig:inte_error}, the bias-corrected integration error exhibits a consistent temporal pattern: a sharp rise at the beginning of the integration window, followed by much slower growth. This pattern is particularly evident for rotation and velocity integration.

\noindent \textbf{Learnable Initial Covariance.}
Standard preintegration propagates the covariance from $\bm{\Sigma}_0 = \mathbf{0}$, which can not capture this error pattern.
AirIMU captures the error pattern in a data-driven manner: its gated recurrent unit (GRU) encoder assigns markedly higher noise to the first samples of the sequence. This compensation, however, is tied to where the recurrent state is initialized. 
In practice, we pre-integrate the IMU data in segments for real-time deployment. 
This, however, force the GRU to be reset at each segment start for the inflated noise to appear, which discards all measurements preceding the segment. 
Running the GRU continuously across segments preserves that context, but the inflated noise then appears only at the start of the whole sequence rather than at each segment start.

To address this issue, we introduce a learnable initial covariance $\bm{\Sigma}_0$, which serves as the starting covariance for the propagation process. 
Moreover, we remove the GRU from the uncertainty prediction branch and instead employ only a CNN encoder shared with the correction prediction. 
Since $\bm{\Sigma}_0$ starts the propagation of every segment, the inflated noise is always present at each segment start, and we don't need to discard the measurements before each segment. 

\noindent \textbf{Metrics-Aware Covariance Supervision.}
The metrics-aware property is enforced during training through a Gaussian negative-log-likelihood loss that supervises the propagated covariance jointly with the state correction~\cite{qiu2023airimu}. Starting the propagation from the learnable $\bm{\Sigma}_0$ and denoting by $\mathbf{e}_r, \mathbf{e}_v, \mathbf{e}_p$ the rotation, velocity, and position integration errors against ground truth, the covariance loss is
\begin{equation}
\label{formula:cov_loss}
L^{\text{cov}} = \!\!\sum_{\star \in \{r, v, p\}}\!\! \frac{1}{2}\Big( \big\| \mathbf{e}_\star \big\|^2_{\bm{\Sigma}_\star} + \ln\det \bm{\Sigma}_\star \Big),
\end{equation}
where $\bm{\Sigma}_r, \bm{\Sigma}_v, \bm{\Sigma}_p$ are the residual covariances obtained from the propagated preintegration covariance. 
The Mahalanobis term rewards small predicted uncertainty on accurate integrations, while the $\ln\det$ term penalizes over-inflated covariance; 
together they drive the predicted uncertainty to track the error magnitude.

\begin{figure}[htbp]
\centering
\includegraphics[width=\columnwidth]{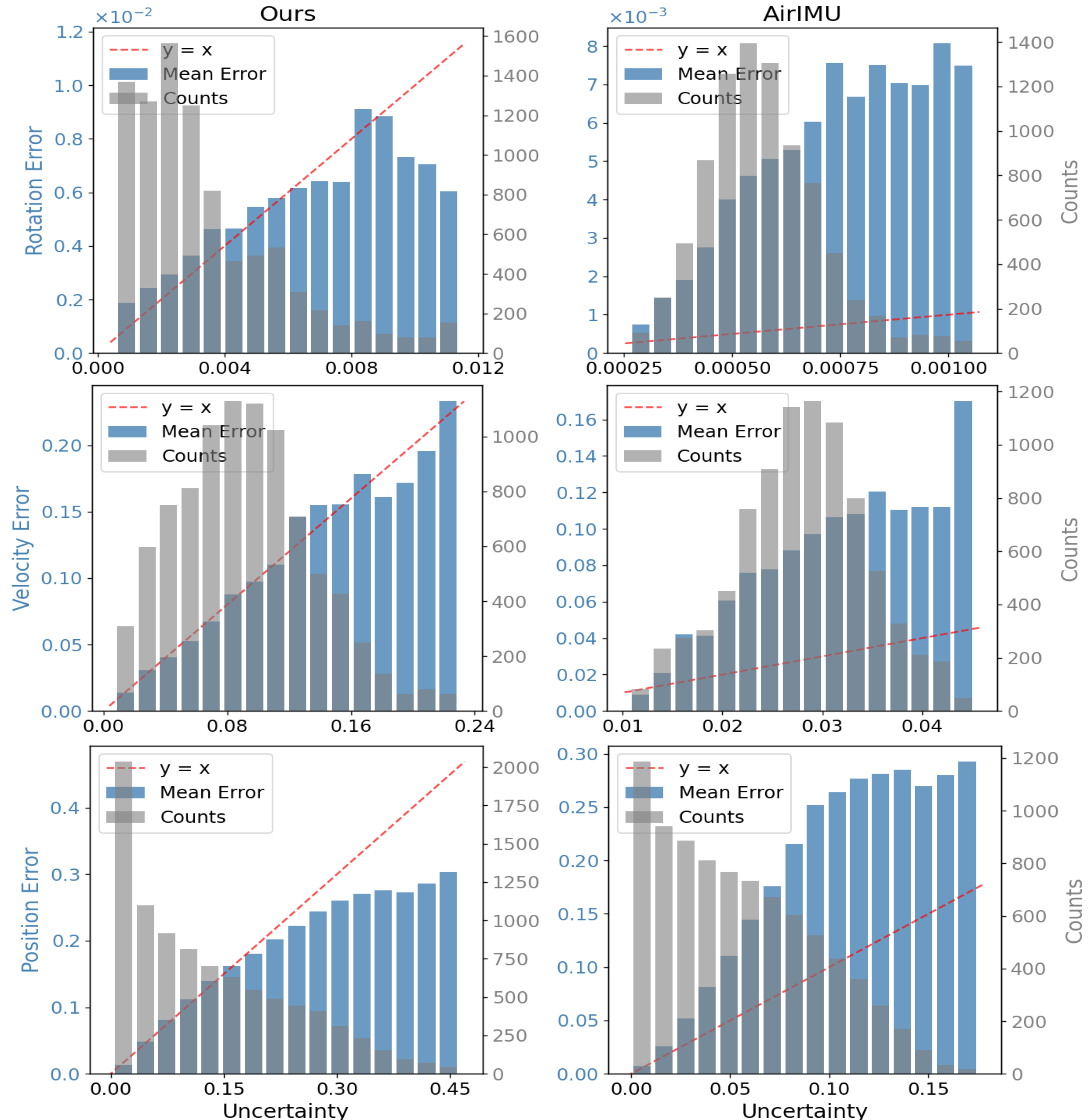}
\caption{Comparison of IMU covariance estimation on unseen EuRoC sequences \texttt{MH\_02}, \texttt{MH\_04}, \texttt{V1\_01}, \texttt{V1\_03}, and \texttt{V2\_02}. Our model produces metrics-aware uncertainties that follow the optimal $y=x$ line, while AirIMU is overconfident. Blue bars denote the mean error in each bin, and grey bars denote the sample counts.}
\label{fig:metrics-aware-imu-cov}
\vspace{-10pt}
\end{figure}

\noindent \textbf{Covariance Fine-tuning.}
On unseen sequences, the predicted covariance tends to be overconfident. To mitigate this, we propose a held-out fine-tuning strategy. Specifically, we split the training set into two subsets, train the full network on the first, and then fine-tune the learnable initial covariance and the uncertainty decoder on the second with all other parameters frozen. As shown in \cref{fig:metrics-aware-imu-cov}, our model produces metrics-aware uncertainties on unseen test sequences, whereas the uncertainties predicted by AirIMU remain overconfident. This metrics-aware covariance is then used in the subsequent VI initialization and calibration.





\subsection{VI Initialization and Calibration}
\label{subsec:init_calib}
\begin{figure}[!t]
\vspace*{6pt}
\centering
\includegraphics[width=\columnwidth]{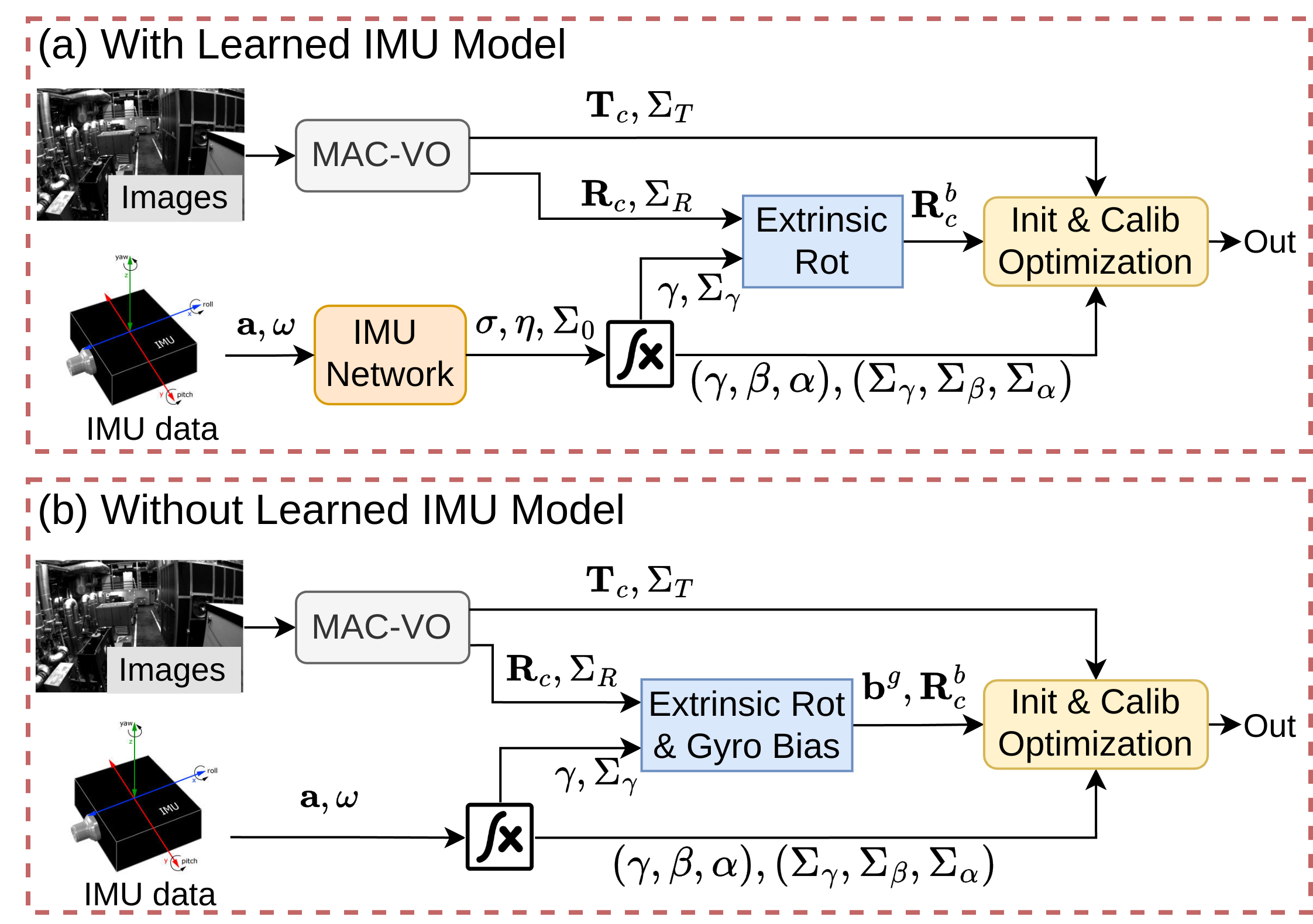}
\caption{Overview of the VI initialization and calibration system built on MAC-I$^2$. (a) With the learned IMU model: corrected IMU measurements and metrics-aware uncertainties. (b) Without the learned IMU model: raw IMU measurements with explicit gyroscope bias estimation.}
\label{fig:system_overview}
\vspace{-10pt}
\end{figure}
As a showcase application, we build a VI initialization and calibration system upon MAC-I$^2$, as illustrated in \cref{fig:system_overview}. The system takes the metrics-aware visual pose covariance (\cref{subsec:visual_cov}) and inertial covariance (\cref{subsec:imu_cov}), and fuses the two modalities in a joint optimization weighted by these covariances. Since the learned IMU model requires training data and may not always be available, we provide two workflows, while the second one illustrates the effectiveness of the metrics-aware visual pose covariance:

\textbf{(a) With Learned IMU Model (\cref{fig:system_overview}a):} The learned IMU network processes raw IMU data to output corrected preintegration terms and predicted uncertainties $(\bm{\eta}^{\text{gyro}}, \bm{\eta}^{\text{acc}})$. Since measurements are already corrected, IMU biases are not explicitly estimated in this workflow.

\textbf{(b) Without Learned IMU Model (\cref{fig:system_overview}b):} Raw IMU data undergoes standard preintegration. Gyroscope bias $\mathbf{b}_g$ is explicitly estimated during initialization, while accelerometer bias is ignored since it is coupled with the gravity vector and cannot be distinguished in a small motion. In addition, IMU uncertainty is pre-calibrated or manually tuned.

\subsection{Gyroscope Bias and Extrinsic Rotation Estimation}
When the learned IMU model is unavailable, we explicitly estimate the gyroscope bias $\mathbf{b}_g$ at this stage. For the calibration task, we also estimate the camera-IMU rotation $\mathbf{q}^c_b$. This can be achieved by minimizing the difference between the rotation from vision and that from IMU integration.
\begin{equation}
\label{formula:gyro_extrinsic}
\begin{aligned}
    \min_{\delta \mathbf{b}_g, \mathbf{q^c_b}} & \sum_{k \in \mathcal{B}} \left\| \text{Log}(\mathbf{q}_c^b \mathbf{q}^{c_{k+1}}_{c_{k}} (\mathbf{q}^b_c)^{-1} \bm{\hat{\gamma}}_{b_{k+1}}^{b_k}) \right\|^2_{\bm{\Sigma}_{r_k}},
    \\
    \bm{\hat{\gamma}}_{b_{k+1}}^{b_k} &= \bm{\gamma}_{b_{k+1}}^{b_k} \text{Exp} \left( \mathbf{J}^{\bm{\gamma}}_{\mathbf{b}_g} \delta \mathbf{b}_g \right),
\end{aligned}
\end{equation}
where $\delta \mathbf{b}_g$ is the increment of the gyroscope bias, initialized to zero so that $\mathbf{b}_g=\delta \mathbf{b}_g$, and $\mathcal{B}$ is the set of all the keyframes. $\boldsymbol{\gamma}_{b_{k+1}}^{b_k}$ is the rotation preintegration of the raw measurements, $\hat{\boldsymbol{\gamma}}_{b_{k+1}}^{b_k}$ is the first-order approximation of the rotation preintegration after gyroscope bias correction \cite{forster_imu}, and $\bm{\Sigma}_{r_k}$ is the covariance of the residual.


\subsection{Learned MAC-enhanced Joint Optimization}
The final stage jointly estimates the state variables $\mathcal {X}$ in \eqref{formula:state} by minimizing a weighted sum of visual and inertial residuals, each weighted by its metrics-aware covariance:
\begin{equation}
\label{formula:tight_optimize}
\begin{aligned}
    \min_{\mathcal {X}} \sum_{k \in \mathcal{B}} \left\| \mathbf{r}^{\text{visual}}_k \right\|^2_{\bm{\Sigma}^{\text{visual}}_k} + \left\| \mathbf{r}^{\text{imu}}_k \right\|^2_{\bm{\Sigma}^{\text{imu}}_k}.
\end{aligned}
\end{equation}

\noindent \textbf{Visual Residual.} Instead of constraining raw features, our visual residual constrains the relative pose, which preserves the feature uncertainty through the derived pose covariance while simplifying the nonlinear optimization, making it well suited to initialization. The visual residual $\mathbf{r}^{\text{visual}}_k$ requires consistency between relative camera motion from visual odometry and motion induced by the estimated states:
\begin{equation}
\label{formula:visual_res}
\begin{aligned}
\mathbf{r}^{\text{visual}}_k &=
\text{Log}(\mathbf{T}^{c_k}_{c_{k+1}} \mathbf{T}^c_b (\mathbf{T}^{c_0}_{b_{k+1}})^{-1}\mathbf{T}^{c_0}_{b_k}\mathbf{T}^b_c).
\end{aligned}
\end{equation}

For the covariance of the visual residual, we consider a left perturbation $\bm{\delta \xi}$ of the estimated pose $\mathbf{T}^{c_k}_{c_{k+1}}$, which gives the perturbed residual $\tilde{\mathbf{r}}^\text{visual}_k = \text{Log}(\text{Exp}(\bm{\delta \xi})\text{Exp}(\mathbf{r}^\text{visual}_k)) \approx \mathbf{r}^\text{visual}_k + \mathbf{J}^{-1}_l(\mathbf{r}^\text{visual}_k)\,\bm{\delta \xi}$, then we have:
\begin{equation}
\label{formula:visual_res_cov}
\begin{aligned}
\frac{\partial{\mathbf{r}^\text{visual}_k}}{\partial{\bm{\delta \xi}}} &= \lim_{\bm{\delta \xi} \to \mathbf{0}} \frac{\tilde{\mathbf{r}}^\text{visual}_k - \mathbf{r}^\text{visual}_k}{\bm{\delta \xi}} = \mathbf{J}^{-1}_l(\mathbf{r}^\text{visual}_k),\\
\bm{\Sigma}^{\text{visual}}_k &= \frac{\partial{\mathbf{r}^\text{visual}_k}}{\partial{\bm{\delta \xi}}} \bm{\Sigma}_{\mathbf{T}^{c_{k}}_{c_{k+1}}} (\frac{\partial{\mathbf{r}^\text{visual}_k}}{\partial{\bm{\delta \xi}}})^{T}\\
&= \mathbf{J}^{-1}_l(\mathbf{r}^\text{visual}_k) \bm{\Sigma}_{\mathbf{T}^{c_{k}}_{c_{k+1}}} \mathbf{J}^{-T}_l(\mathbf{r}^\text{visual}_k),
\end{aligned}
\end{equation}
where $\bm{\Sigma}_{\mathbf{T}^{c_{k}}_{c_{k+1}}}$ is the visual pose covariance from \eqref{pose_cov} and $\mathbf{J}^{-1}_l(\mathbf{r}^{\text{visual}}_k)$ is the inverse left Jacobian of $\operatorname{SE}(3)$ at $\mathbf{r}^{\text{visual}}_k$.

\begin{table*}[htbp]
\vspace*{8pt}
\centering
\captionsetup{font=small}
\caption{Initialization errors and success rate (SR) for the 10KFs setting in EuRoC dataset.}
\vspace{-2mm}
\centering
\fontsize{14}{16}\selectfont
\setlength{\tabcolsep}{3pt}
\setlength{\tabcolsep}{3pt}
\renewcommand{\arraystretch}{1}
\resizebox{1\linewidth}{!}{
\begin{tabular}{
l ccccccccccccc 
}
\toprule
 Trajectory & \multicolumn{2}{c}{{MH02}} & \multicolumn{2}{c}{{MH04}} & \multicolumn{2}{c}{{V101}} & \multicolumn{2}{c}{V103} & \multicolumn{2}{c}{V202} & \multicolumn{3}{c}{Avg.} \\
\cmidrule{2-14}
   & Vel(m/s) & G.Dir($\degree$) & Vel(m/s) & G.Dir($\degree$) & Vel(m/s) & G.Dir($\degree$) & Vel(m/s) & G.Dir($\degree$) & Vel(m/s) & G.Dir($\degree$) & Vel(m/s) & G.Dir($\degree$) & SR $\uparrow$ \\
\midrule
\monoicon~VINS-Mono   & 0.085 & 1.093 & 0.367 & 1.447 & 0.107 & 1.582 & 0.187 & 2.748 & 0.100 & 1.182 & 0.169 & 1.611 &  \textcolor{c2}{\textbf{0.318}} \\
\monoicon~DRT-t      & 0.104 & 0.962 & 0.261 & 1.159 & 0.080 & 0.966 & 0.145 & 1.679 & 0.081 & \underline{0.918} & 0.134 & 1.137 & \textcolor{c5}{\textbf{0.633}} \\
\monoicon~DRT-l      & 0.065 & 0.967 & 0.189 & 1.091 & 0.057 & 1.018 & 0.269 & 2.235 & 0.109 & 1.206 & 0.138 & 1.303 & \textcolor{c6}{\textbf{0.640}} \\
\monoicon~sqrt-VINS mono$^{\star}$   & 0.059 & 1.792 & 0.121 & 0.857 & 0.048 & 0.841 & 0.075 & 1.119 & 0.062 & 0.753 & 0.073 & 1.072 &  \textcolor{c7}{\textbf{0.704}} \\
\monoicon~FF-VIO-Init$^{\star}$   & 0.054 & 0.934 & 0.139 & 0.966 & 0.025 & 0.884 & 0.032 & 1.045 & 0.028 & 0.740 & 0.055 & 0.914 &  \textcolor{c8}{\textbf{0.866}} \\
\midrule
\stereoicon~VINS-Fusion  & 0.023 & 2.191 & 0.050 & 1.905 & 0.019 & 1.943 & \underline{0.041} & 2.467 & \underline{0.024} & 2.080 & \underline{0.031} & 2.117 &  \textcolor{c4}{\textbf{0.582}} \\
\stereoicon~ORB-SLAM 3          & \underline{0.011} & 0.928 & 0.055 & 1.231 & 0.011 & 0.876 & 0.136 & 2.703 & 0.038 & 1.218 & 0.050 & 1.391 &  \textcolor{c8}{\textbf{0.820}}\\
\stereoicon~Stereo-NEC  & \underline{0.011} & 0.946 & 0.038 & 1.363 & \textbf{0.008} & 0.926 & 0.101 & 2.946 & 0.026 & 1.404 & 0.037 & 1.517 &  \textcolor{c8}{\textbf{0.823}} \\
\stereoicon~sqrt-VINS stereo$^{\star}$ & 0.046 & 1.041 & 0.102 & 0.839 & 0.031 & 1.500 & 0.070 & 1.063 & 0.062 & 0.800 & 0.062 & 1.049 &  \textcolor{c7}{\textbf{0.736}} \\
\stereoicon~\textbf{MAC-I$^2$ w/o Learned IMU}                  & \underline{0.011} & \underline{0.820} & \textbf{0.022} & \underline{0.800} & \underline{0.010} & \underline{0.872} &\textbf{0.025} & \underline{1.149} & \textbf{0.020} & \textbf{0.804} & \textbf{0.018} & \underline{0.889} &  \textcolor{c11}{\textbf{0.999}}\\
\stereoicon~\textbf{MAC-I$^2$ w/ Learned IMU}                  & \textbf{0.010} & \textbf{0.166} & \underline{0.023} & \textbf{0.208} & \underline{0.010} & \textbf{0.332} & \textbf{0.025} & \textbf{0.581} & \underline{0.024} & \textbf{0.804} & \textbf{0.018} & \textbf{0.418} & \textcolor{c11}{\textbf{0.999}}\\
\bottomrule
\multicolumn{14}{l}{
\quad
\monoicon~\hspace{.1mm} Monocular method.
\quad
\stereoicon~\hspace{.1mm} Stereo method.
\quad
$^{\star}$~With good initial guess of IMU biases; performance degrades otherwise. Excluded from ranking.
} \\
\end{tabular}
}
\vspace{-5pt}
\label{tab:EuRoCEven}
\end{table*}

\noindent \textbf{IMU Residual.} The IMU residual $\mathbf{r}^{\text{imu}}_k = \begin{bmatrix}
\mathbf{r}_{\bm{\gamma}_k}^T, \mathbf{r}_{\bm{\beta}_k}^T, \mathbf{r}_{\bm{\alpha}_k}^T
\end{bmatrix}^T$ is derived from \eqref{formula:imu_integration}:
{\small
\begin{align}
\label{formula:inte_res_r}
\mathbf{r}_{\bm{\gamma}_k} &= \text{Log} (\mathbf{R}^{b_{k+1}}_{c_0} \mathbf{R}^{c_0}_{b_k}\tilde{\boldsymbol{\gamma}}_{b_{k+1}}^{b_k}), \\
\label{formula:inte_res_v}
\mathbf{r}_{\bm{\beta}_k} &= \mathbf{R}^{b_k}_{c_0} \mathbf{R}^{c_0}_{b_{k+1}} \mathbf{v}_{b_{k+1}}^{b_{k+1}} \!-\! \mathbf{v}_{b_k}^{b_k} \!-\! \mathbf{R}^{b_k}_{c_0} \mathbf{g}^{c_0} \Delta t_{kk+1} \!-\! \tilde{\boldsymbol{\beta}}_{b_{k+1}}^{b_k},
\\
\label{formula:inte_res_p}
\mathbf{r}_{\bm{\alpha}_k} &= \mathbf{R}^{b_k}_{c_0} \mathbf{p}_{b_{k+1}}^{c_0} \!-\! \mathbf{R}^{b_k}_{c_0} \mathbf{p}_{b_k}^{c_0} \!-\! \mathbf{v}_{b_k}^{b_k}\Delta t_{kk+1} \\ \notag \!&\quad -\! \tfrac{1}{2}\mathbf{R}^{b_k}_{c_0}\mathbf{g}^{c_0}\Delta t_{kk+1}^2 \!-\! \tilde{\boldsymbol{\alpha}}_{b_{k+1}}^{b_k},
\end{align}}
where $\tilde{\boldsymbol{\gamma}}_{b_{k+1}}^{b_k}, \tilde{\boldsymbol{\beta}}_{b_{k+1}}^{b_k}, \tilde{\boldsymbol{\alpha}}_{b_{k+1}}^{b_k}$ denote the preintegration terms used in each workflow. With the learned IMU model, these are the network-corrected preintegrations and explicit bias estimation is unnecessary. Without the learned IMU model, these reduce to first-order bias-corrected terms in standard IMU preintegration~\cite{forster_imu}. The IMU residual covariance can be obtained from the predicted preintegration covariance.

%

For all residuals above, we derive analytical Jacobians on the $\operatorname{SO}(3)$/$\operatorname{SE}(3)$ manifolds for efficient optimization~\cite{wang2023pypose}.

\section{Experiments}
\label{sec:experiments}

\subsection{Experiment Setup}

\noindent\textbf{Datasets.} 
We evaluate initialization on public benchmarks: EuRoC~\cite{burri2016euroc} and VBR~\cite{brizi2024vbr}. EuRoC is an MAV dataset with diverse motions; VBR covers large-scale and dynamic scenarios, from which we use the 8 sequences with ground-truth poses. For camera-IMU extrinsic calibration, we use the 4 IMU calibration sequences from TUM-VI~\cite{schubert2018tum}, which contain rapid motions exciting all six degrees of freedom.

\noindent\textbf{Baselines.}
For initialization, we compare against several state-of-the-art geometric methods. \textit{VINS-Mono}~\cite{qin2018vins} estimates gravity and velocity in a loosely-coupled manner, whereas its stereo counterpart \textit{VINS-Fusion}~\cite{qin2019b} optimizes tightly after gyroscope bias estimation. \textit{ORB-SLAM3}~\cite{campos2021orb} uses an inertial-only initialization. \textit{DRT}~\cite{he2023rotation} decouples rotation and translation and estimates gyroscope bias with the Normal Epipolar Constraint (NEC); we evaluate both its tightly-coupled (\textit{DRT-t}) and loosely-coupled (\textit{DRT-l}) variants. \textit{Stereo-NEC}~\cite{wang2024stereo} integrates NEC into \textit{ORB-SLAM3}. \textit{sqrt-VINS}~\cite{peng2025sqrtvins} recovers the minimal states in closed form without triangulation and refines them with a square-root filter, in both monocular and stereo configurations. \textit{FF-VIO-Init}~\cite{zhang2026efficient} eliminates explicit feature tracking and formulates the initialization problem using constraints derived from up-to-scale point clouds predicted by a feed-forward 3D model.
For calibration, we compare with \textit{VINS-Mono}, which provides a complete calibration solution, and use \textit{Kalibr}\cite{kalibr} for ground truth.

\noindent\textbf{Evaluation Metrics.}
For VI initialization, we consider:
\begin{itemize}
\item Gravity direction error (G.Dir): $\arccos(\mathbf{g}^{c_0} \cdot \tilde{\mathbf{g}}^{c_0})$, where $\mathbf{g}^{c_0}$, $\tilde{\mathbf{g}}^{c_0}$ denote the estimated and ground truth gravity.
\item Velocity RMSE: $\|\mathbf{v}^{b_k}_{b_k}-\tilde{\mathbf{v}}^{b_k}_{b_k}\| / N$, where $\mathbf{v}^{b_k}_{b_k}$, $\tilde{\mathbf{v}}^{b_k}_{b_k}$ denote the estimated and ground truth body velocity of frame $k$, $N$ is the total number of keyframes.
\item Gyroscope bias error: following \cite{he2023rotation, mu2025robust}, the error is the relative deviation $\left| \|\mathbf{b}_g\| - \|\bar{\mathbf{b}}^g\| \right| / \|\bar{\mathbf{b}}^g\|$, where $\bar{\mathbf{b}}^g$ is the mean bias along the ground-truth trajectory.
\item Success rate (SR): the fraction of segments solved with G.Dir error $<2^\circ$ and velocity RMSE $<0.1$~m/s on EuRoC, $<0.3$~m/s on VBR.
\end{itemize}
For calibration, we evaluate the estimated extrinsics with:
\begin{itemize}
\item Rotation error: $\|\text{Log}((\mathbf{q}^b_c)^{-1} * \tilde{\mathbf{q}}^b_c)\| \cdot \frac{180}{\pi}$, where $\mathbf{q}^b_c$, $\tilde{\mathbf{q}}^b_c$ denote the estimated and ground truth extrinsic rotation.
\item Translation error: $\|\mathbf{p}^b_c - \tilde{\mathbf{p}}^b_c\|$, where $\mathbf{p}^b_c$, $\tilde{\mathbf{p}}^b_c$ denote the estimated and ground truth extrinsic translation.
\end{itemize}

\noindent\textbf{Training Details.}
The visual model is trained on the synthetic TartanAir dataset~\cite{wang2020tartanair}. On EuRoC, the IMU model is trained on \texttt{MH\_01}, \texttt{MH\_05}, \texttt{V2\_01}, and \texttt{V2\_03}, with uncertainty fine-tuned on \texttt{MH\_03} and \texttt{V1\_02}.

\begin{figure}[!t]
\vspace*{7pt}
\centering
\includegraphics[width=\columnwidth]{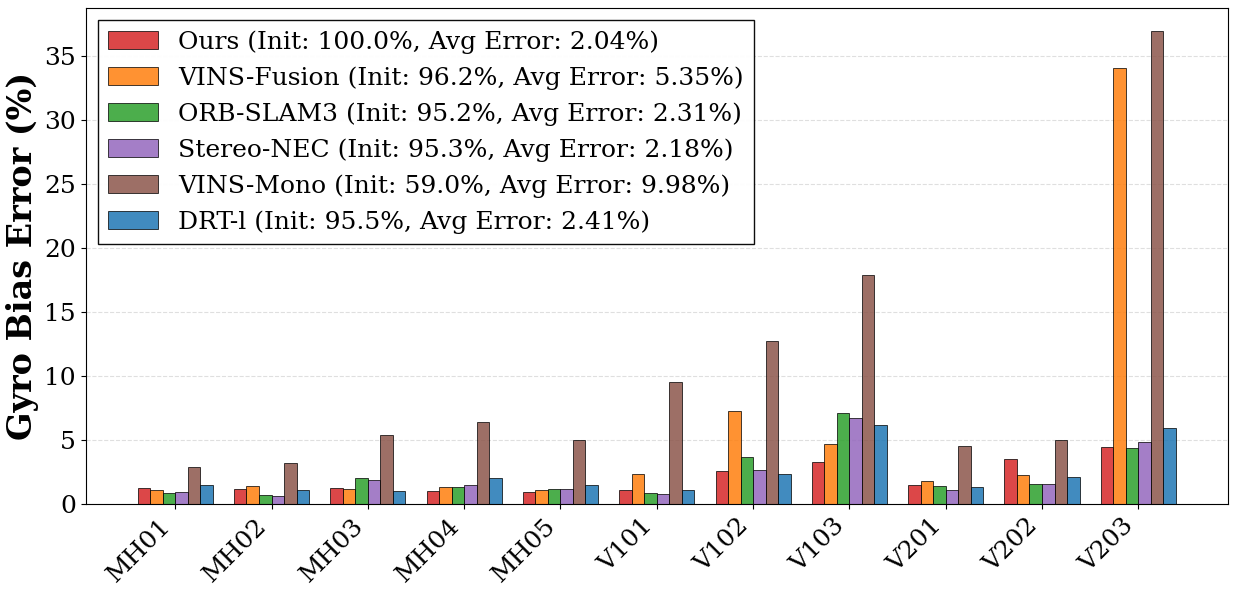}
\caption{Comparison of gyroscope bias magnitude estimation on all the sequences of EuRoC dataset.}
\label{fig:gyro_bias}
\vspace{-10pt}
\end{figure}

\begin{table*}[htbp]
\vspace*{8pt}
\centering
\captionsetup{font=small}
\caption{Detailed initialization results for the 10KFs setting in VBR dataset. Vel is in m/s, G.Dir in degrees.}
\vspace{-2mm}
    \resizebox{1\linewidth}{!}{
    \fontsize{14}{16}\selectfont
    \setlength{\tabcolsep}{3pt}
\begin{tabular}{l c@{\hspace{8pt}}c@{\hspace{13pt}} c@{\hspace{8pt}}c@{\hspace{13pt}} c@{\hspace{8pt}}c@{\hspace{13pt}} c@{\hspace{8pt}}c@{\hspace{13pt}} c@{\hspace{8pt}}c@{\hspace{13pt}} c@{\hspace{8pt}}c@{\hspace{13pt}} c@{\hspace{8pt}}c@{\hspace{13pt}} c@{\hspace{8pt}}c@{\hspace{13pt}} c@{\hspace{8pt}}c@{\hspace{8pt}}c}
\toprule
 Trajectory
 & \multicolumn{2}{c@{\hspace{13pt}}}{cam0}
 & \multicolumn{2}{c@{\hspace{13pt}}}{cam1}
 & \multicolumn{2}{c@{\hspace{13pt}}}{cia0}
 & \multicolumn{2}{c@{\hspace{13pt}}}{cia1}
 & \multicolumn{2}{c@{\hspace{13pt}}}{col0}
 & \multicolumn{2}{c@{\hspace{13pt}}}{dia0}
 & \multicolumn{2}{c@{\hspace{13pt}}}{pin0}
 & \multicolumn{2}{c@{\hspace{13pt}}}{spa0}
 & \multicolumn{3}{c}{Avg.} \\
\cmidrule{2-20}
 & Vel & G.Dir & Vel & G.Dir & Vel & G.Dir & Vel & G.Dir & Vel & G.Dir & Vel & G.Dir & Vel & G.Dir & Vel & G.Dir & Vel & G.Dir & SR $\uparrow$ \\
\midrule
\monoicon~VINS-Mono  & 2.149 & 2.077 & 2.763 & 2.222 & 4.474 & 2.052 & 3.008 & 1.730 & 1.084 & \textbf{1.055} & 0.426 & \underline{0.925} & 0.514 & \textbf{0.903} & 0.522 & 0.913 & 1.771 & 1.457 & \textcolor{c5}{\textbf{0.080}} \\
\monoicon~sqrt-VINS mono  & 2.822 & 1.792 & 4.328 & 2.039 & 4.363 & 2.446 & 3.919 & 2.054 & 0.772 & \underline{1.267} & 0.279 & 2.644 & -     & -     & 0.812 & \textbf{0.795} & 3.423 & 1.991 & \textcolor{c1}{\textbf{0.000}} \\
\midrule
\stereoicon~VINS-Fusion  & 1.333 & 2.725 & 2.126 & 2.931 & 2.340 & 4.106 & 1.958 & 3.069 & 0.571 & 1.617 & 0.245 & 1.376 & 0.182 & 1.289 & 0.235 & 1.184 & 1.119 & 2.272 & \textcolor{c6}{\textbf{0.369}} \\
\stereoicon~ORB-SLAM 3  & \underline{0.303} & \underline{1.561} & \underline{0.409} & \underline{1.152} & 0.541 & 1.406 & \underline{0.420} & \underline{0.985} & \underline{0.240} & 1.803 & \underline{0.134} & 1.232 & \underline{0.072} & \underline{1.010} & \underline{0.110} & 0.862 & 0.280 & \underline{1.251} & \textcolor{c8}{\textbf{0.561}} \\
\stereoicon~Stereo-NEC  & 0.304 & 1.814 & 0.416 & 1.528 & \underline{0.520} & 1.545 & 0.433 & 1.353 & \textbf{0.217} & 2.047 & \textbf{0.128} & 1.933 & \textbf{0.068} & 1.232 & \textbf{0.107} & 1.043 & \underline{0.276} & 1.556 & \textcolor{c7}{\textbf{0.514}} \\
\stereoicon~sqrt-VINS stereo  & 1.559 & 1.710 & 1.758 & 1.186 & 2.266 & \underline{1.130} & 2.302 & 1.318 & 0.701 & 1.587 & 0.246 & \textbf{0.725} & -     & -     & 0.584 & \underline{0.815} & 1.722 & 1.363 & \textcolor{c3}{\textbf{0.030}} \\
\stereoicon~\textbf{MAC-I$^2$ w/o Learned IMU}  & \textbf{0.135} & \textbf{1.521} & \textbf{0.215} & \textbf{1.105} & \textbf{0.165} & \textbf{0.959} & \textbf{0.126} & \textbf{0.653} & 0.272 & 1.348 & 0.186 & 1.184 & 0.128 & 1.111 & 0.296 & 1.194 & \textbf{0.191} & \textbf{1.136} & \textcolor{c11}{\textbf{0.803}} \\
\bottomrule
\multicolumn{20}{l}{
\quad
\monoicon~\hspace{.1mm} Monocular method.
\quad
\stereoicon~\hspace{.1mm} Stereo method.
\quad
- The initialization fails on all segments.}\\
\end{tabular}
}
\vspace{-5pt}
\label{tab:VBR}
\end{table*}

\subsection{Initialization Accuracy and Robustness Evaluation}

Each sequence is divided into 2.5-second segments, and initialization is performed on each segment with 10 keyframes. For our method with the learned IMU model, evaluation is restricted to sequences not used in training.

On EuRoC (Table~\ref{tab:EuRoCEven}), \textit{MAC-I$^2$ w/ Learned IMU} attains an average gravity error of 0.418$\degree$ and velocity RMSE of 0.018 m/s, improving over the best baseline on each metric, \textit{DRT-t} (1.137$\degree$) and \textit{VINS-Fusion} (0.031 m/s), by about 63\% and 42\%. Even without the learned IMU model, our method reaches 0.889$\degree$ and 0.018 m/s, still outperforming all baselines. More notably, our method achieves a 0.999 success rate across all segments. The success rate gap is pronounced in challenging sequences: on V103, \textit{VINS-Mono}, \textit{DRT-l}, and \textit{VINS-Fusion} achieve success rates of only 0.040, 0.279, and 0.408, and even the strongest baseline, \textit{Stereo-NEC}, drop to 0.537, while our method achieves 1.00. In addition, \textit{sqrt-VINS} and \textit{FF-VIO-Init} rely on a good initial guess of the IMU biases and degrade otherwise, so we report their results under a good bias initialization. We also evaluate gyroscope bias as it is critical to VI initialization~\cite{he2023rotation}. Fig.~\ref{fig:gyro_bias} shows that our method estimates the gyroscope bias accurately and robustly across all sequences.

On VBR (Table~\ref{tab:VBR}), baselines largely fail: \textit{VINS-Mono} and \textit{sqrt-VINS} fall below a 0.10 success rate, and even the best baseline \textit{ORB-SLAM3} reaches only 0.561. In contrast, our method reaches 0.803, and attains the lowest average errors of 1.136$\degree$ and 0.191 m/s, demonstrating strong robustness.

\subsection{Calibration Results}
\begin{figure}[!t]
\centering
\includegraphics[width=\columnwidth]{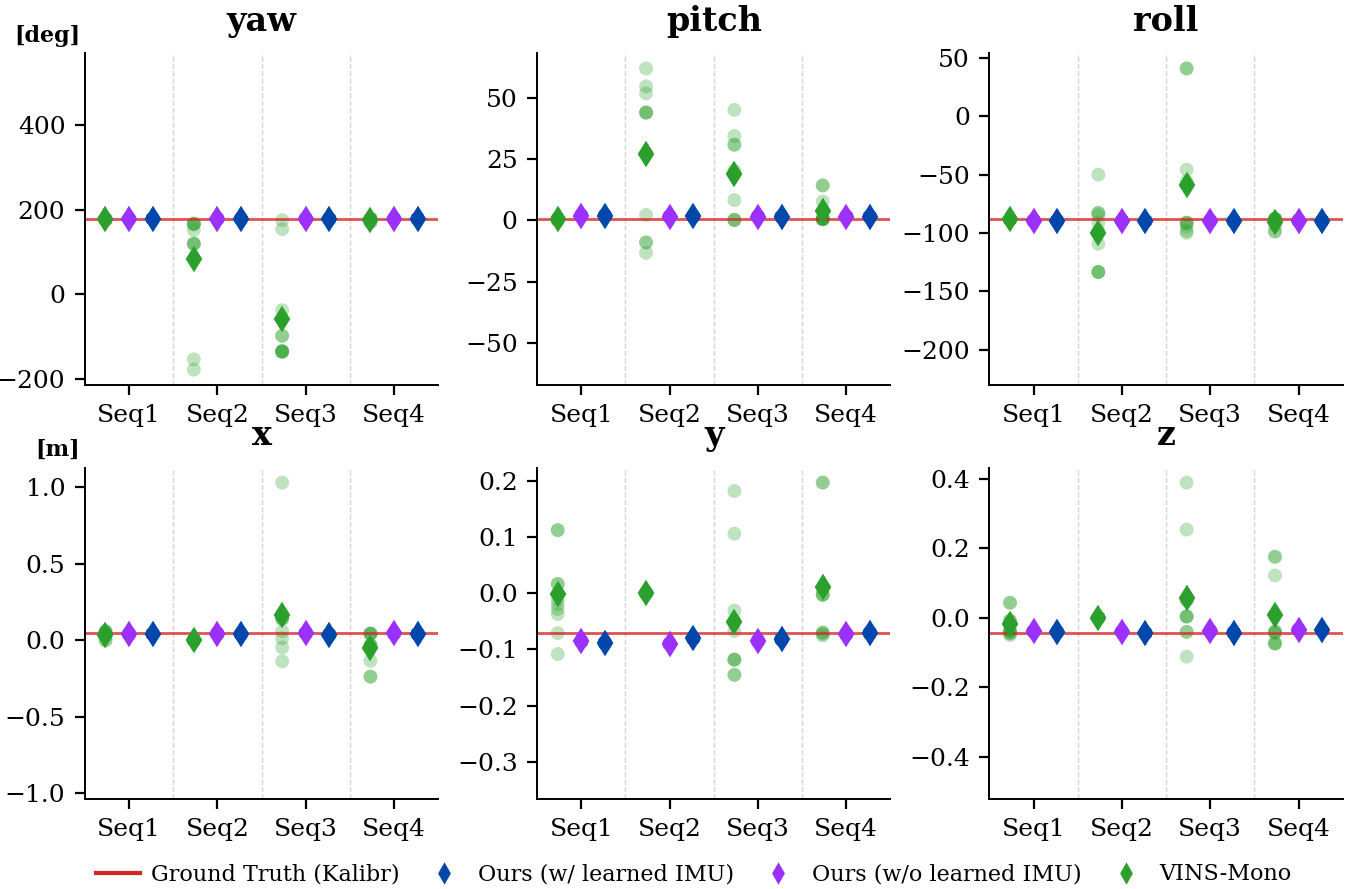}
\caption{Comparison of camera-IMU extrinsic calibration on TUM-VI calib-imu sequences.}
\vspace{-7pt}
\label{fig:calib1}
\end{figure}

We evaluate calibration on the TUM-VI IMU calibration sequences, which contain aggressive six-DoF motions, repeating each experiment 10 times per sequence. As shown in Fig.~\ref{fig:calib1}, our method produces accurate and consistent calibration across all trials, whereas \textit{VINS-Mono} loses tracking in most trials and drifts substantially. This is because geometric feature-based methods are challenged by the highly aggressive motion patterns in the TUM-VI calibration sequences, which were originally designed for checkerboard-based calibration. Under such conditions, our method remains robust and is able to deliver stable and accurate calibration results.

\begin{table}[!t]
    \caption{Initialization errors on the EuRoC dataset under different covariance settings. 
}

    \label{cov_euroc}
    \centering
    \resizebox{1\linewidth}{!}{
    \fontsize{14}{16}\selectfont
    \setlength{\tabcolsep}{3pt}
    \begin{tabular}{c@{\hspace{1.5pt}}|r@{\hspace{1pt}}c@{\hspace{2.5pt}}r| cccccc}
        \toprule
        \textbf{Metrics} & 
        \multicolumn{3}{c}{Methods} 
        & {MH02} & {MH04} & {V101} & {V103} & {V202} & {Avg} \\
        \midrule

\multirow{4}{*}{\shortstack{\textbf{Velocity}\\\textbf{(m/s)}}}

        & Id Vis & + & Id IMU
        & \cellcolor{g3} 0.013 & \cellcolor{g3} 0.031 & \cellcolor{g1} \textbf{0.010} & \cellcolor{g3} 0.026 & \cellcolor{g1} \textbf{0.022} & \cellcolor{g3} 0.021 \\

        & Id Vis & + & Learned IMU
        & \cellcolor{g4} 0.040 & \cellcolor{g4} 0.073 & \cellcolor{g4} 0.037 & \cellcolor{g4} 0.061 & \cellcolor{g4} 0.056 & \cellcolor{g4} 0.053 \\
           & Learned Vis & + & Id IMU
        & \cellcolor{g2} 0.011 & \cellcolor{g1} \textbf{0.021} & \cellcolor{g3} 0.011 & \cellcolor{g2} 0.027 & \cellcolor{g3} 0.029 & \cellcolor{g2} 0.020 \\
         & Learned Vis & + & Learned IMU
        & \cellcolor{g1} \textbf{0.010} & \cellcolor{g2} 0.023 & \cellcolor{g1} \textbf{0.010} & \cellcolor{g1} \textbf{0.025} & \cellcolor{g2} 0.024 & \cellcolor{g1} \textbf{0.018} \\
        \midrule
\multirow{4}{*}{\shortstack{\textbf{Gravity}\\\textbf{Direction (deg)}}}

        & Id Vis & + & Id IMU
        & \cellcolor{g3} 0.239 & \cellcolor{g3} 0.284 & \cellcolor{g3} 0.353 & \cellcolor{g3} 0.671 & \cellcolor{g1} \textbf{0.766} & \cellcolor{g2} 0.463 \\

        & Id Vis & + & Learned IMU
        & \cellcolor{g4} 0.284 & \cellcolor{g4} 0.459 & \cellcolor{g4} 0.385 & \cellcolor{g2} 0.655 & \cellcolor{g4} 0.865 & \cellcolor{g4} 0.530 \\
 & Learned Vis & + & Id IMU
        & \cellcolor{g2} 0.231 & \cellcolor{g2} 0.248 & \cellcolor{g2} 0.346 & \cellcolor{g4} 0.768 & \cellcolor{g3} 0.810 & \cellcolor{g3} 0.481 \\
        & Learned Vis & + & Learned IMU
        & \cellcolor{g1} \textbf{0.166} & \cellcolor{g1} \textbf{0.208} & \cellcolor{g1} \textbf{0.332} & \cellcolor{g1} \textbf{0.581} & \cellcolor{g2} 0.804 & \cellcolor{g1} \textbf{0.418} \\
        \midrule

\multirow{4}{*}{\shortstack{\textbf{Body}\\\textbf{Rotation (deg)}}}

        & Id Vis & + & Id IMU
        & \cellcolor{g3} 0.193 & \cellcolor{g4} 0.264 & \cellcolor{g4} 0.224 & \cellcolor{g4} 0.597 & \cellcolor{g4} 0.534 & \cellcolor{g4} 0.362 \\

        & Id Vis & + & Learned IMU
        & \cellcolor{g2} 0.107 & \cellcolor{g2} 0.106 & \cellcolor{g2} 0.192 & \cellcolor{g2} 0.365 & \cellcolor{g2} 0.428 & \cellcolor{g2} 0.240 \\
          & Learned Vis & + & Id IMU
        & \cellcolor{g4} 0.215 & \cellcolor{g3} 0.193 & \cellcolor{g3} 0.193 & \cellcolor{g3} 0.544 & \cellcolor{g3} 0.457 & \cellcolor{g3} 0.320 \\
          & Learned Vis & + & Learned IMU
        & \cellcolor{g1} \textbf{0.084} & \cellcolor{g1} \textbf{0.099} & \cellcolor{g1} \textbf{0.161} & \cellcolor{g1} \textbf{0.362} & \cellcolor{g1} \textbf{0.368} & \cellcolor{g1} \textbf{0.215} \\
        \midrule
\multirow{4}{*}{\shortstack{\textbf{Body}\\\textbf{Translation (m)}}}

        & Id Vis & + & Id IMU
        & \cellcolor{g1} \textbf{0.009} & \cellcolor{g3} 0.034 & \cellcolor{g1} \textbf{0.008} & \cellcolor{g2} 0.043 & \cellcolor{g2} 0.024 & \cellcolor{g2} 0.024 \\
     
        & Id Vis & + & Learned IMU
        & \cellcolor{g4} 0.024 & \cellcolor{g4} 0.048 & \cellcolor{g3} 0.026 & \cellcolor{g4} 0.055 & \cellcolor{g4} 0.042 & \cellcolor{g3} 0.039 \\
         
        & Learned Vis & + & Id IMU
        & \cellcolor{g2} 0.010 & \cellcolor{g2} 0.033 & \cellcolor{g1} \textbf{0.008} & \cellcolor{g2} 0.043 & \cellcolor{g2} 0.024 & \cellcolor{g2} 0.024 \\
          & Learned Vis & + & Learned IMU
        & \cellcolor{g1} \textbf{0.009} & \cellcolor{g1} \textbf{0.031} & \cellcolor{g1} \textbf{0.008} & \cellcolor{g1} \textbf{0.037} & \cellcolor{g1} \textbf{0.022} & \cellcolor{g1} \textbf{0.021} \\
        \bottomrule

    \end{tabular}
    }
    \vspace{-5pt}
\end{table}

\begin{table}[!t]
   \caption{Initialization errors on the EuRoC dataset with and without the learned IMU model, and against AirIMU.}
    \label{airimu_euroc}
    \centering
    \fontsize{15}{16}\selectfont
    \resizebox{1\linewidth}{!}{
\begin{tabular}{c|l|cccccc}
    \toprule
    \textbf{Metrics} & {Methods} 
    & {MH02} & {MH04} & {V101} & {V103} & {V202} & {Avg} \\
    \midrule
    \multirow{3}{*}{\shortstack{\textbf{Velocity}\\ \textbf{(m/s)}}}
    & w/o Learned IMU 
    & \cellcolor{g2} 0.011 & \cellcolor{g2} 0.022 & \cellcolor{g2} 0.010 & \cellcolor{g1} \textbf{0.025} & \cellcolor{g1} \textbf{0.020} & \cellcolor{g1} \textbf{0.017} \\
    & AirIMU
    & \cellcolor{g4} 0.013 & \cellcolor{g4} 0.031 & \cellcolor{g1} \textbf{0.008} & \cellcolor{g4} 0.028 & \cellcolor{g1} \textbf{0.020} & \cellcolor{g4} 0.020 \\
    & Our Learned IMU
    & \cellcolor{g1} \textbf{0.010} & \cellcolor{g1} \textbf{0.023} & \cellcolor{g2} 0.010 & \cellcolor{g1} \textbf{0.025} & \cellcolor{g4} 0.024 & \cellcolor{g2} 0.018 \\
    \midrule
    \multirow{3}{*}{\shortstack{\textbf{Gravity }\\ \textbf{Direction (deg)}}}
    & w/o Learned IMU 
    & \cellcolor{g4} 0.820 & \cellcolor{g4} 0.800 & \cellcolor{g4} 0.872 & \cellcolor{g4} 1.149 & \cellcolor{g2} 0.804 & \cellcolor{g4} 0.889 \\
    & AirIMU
    & \cellcolor{g2} 0.442 & \cellcolor{g2} 0.438 & \cellcolor{g2} 0.589 & \cellcolor{g2} 0.794 & \cellcolor{g4} 0.791 & \cellcolor{g2} 0.611 \\
    & Our Learned IMU
    & \cellcolor{g1} \textbf{0.166} & \cellcolor{g1} \textbf{0.208} & \cellcolor{g1} \textbf{0.332} & \cellcolor{g1} \textbf{0.581} & \cellcolor{g1} \textbf{0.804} & \cellcolor{g1} \textbf{0.418} \\
    \midrule
    \multirow{3}{*}{\shortstack{\textbf{Body}\\ \textbf{Rotation (deg)}}}
    & w/o Learned IMU 
    & \cellcolor{g4} 0.140 & \cellcolor{g4} 0.149 & \cellcolor{g4} 0.185 & \cellcolor{g2} 0.420 & \cellcolor{g2} 0.398 & \cellcolor{g4} 0.258 \\
    & AirIMU
    & \cellcolor{g2} 0.115 & \cellcolor{g2} 0.144 & \cellcolor{g2} 0.174 & \cellcolor{g4} 0.428 & \cellcolor{g4} 0.409 & \cellcolor{g2} 0.254 \\
    & Our Learned IMU
    & \cellcolor{g1} \textbf{0.084} & \cellcolor{g1} \textbf{0.099} & \cellcolor{g1} \textbf{0.161} & \cellcolor{g1} \textbf{0.362} & \cellcolor{g1} \textbf{0.368} & \cellcolor{g1} \textbf{0.215} \\
    \midrule
    \multirow{3}{*}{\shortstack{\textbf{Body}\\\textbf{Translation (m)}}}
    & w/o Learned IMU 
    & \cellcolor{g1} \textbf{0.009} & \cellcolor{g2} 0.031 & \cellcolor{g1} \textbf{0.008} & \cellcolor{g2} 0.038 & \cellcolor{g1} \textbf{0.021} & \cellcolor{g2} 0.022 \\
    & AirIMU
    & \cellcolor{g1} \textbf{0.009} & \cellcolor{g4} 0.033 & \cellcolor{g1} \textbf{0.008} & \cellcolor{g2} 0.038 & \cellcolor{g1} \textbf{0.021} & \cellcolor{g2} 0.022 \\
    & Our Learned IMU
    & \cellcolor{g1} \textbf{0.009} & \cellcolor{g1} \textbf{0.031} & \cellcolor{g1} \textbf{0.008} & \cellcolor{g1} \textbf{0.037} & \cellcolor{g4} 0.022 & \cellcolor{g1} \textbf{0.021} \\
    \bottomrule
\end{tabular}
    \vspace{-5pt}
    }
    \label{tab:learned_imu_ablation}
\end{table}

\section{Ablation Study}

\noindent\textbf{Effectiveness of the Learned Covariance:}
We ablate the learned visual and IMU covariances with four configurations: (i) Learned Vis + Learned IMU (our full model), (ii) Id Vis + Id IMU, (iii) Learned Vis + Id IMU, and (iv) Id Vis + Learned IMU, where ``Id'' denotes an identity covariance. We evaluate the initial gravity, velocity, and pose (Table~\ref{cov_euroc}).

The results show that our learned covariance models consistently improve initialization accuracy across all metrics and sequences. 
The gains are most notable in initial pose estimation, where body rotation error drops from 0.362$\degree$ to 0.215$\degree$ (40.6\% $\downarrow$) and body translation error from 0.024 m to 0.021 m (12.5\% $\downarrow$).
The learned visual covariance also benefits gyroscope bias estimation, reducing the average bias magnitude error from 3.9\% to 2.0\% (48.7\% $\downarrow$).

\noindent\textbf{Effectiveness of the Learned IMU Model:}
Table~\ref{tab:learned_imu_ablation} ablates the effect of the learned IMU model. Incorporating it yields consistent and significant improvements in orientation-related metrics, including gravity direction and body rotation errors, across all sequences. These gains stem not only from the learned IMU covariance, but also from the model's correction of accelerometer measurements, which directly affects gravity alignment and long-term integration.

We further compare against using AirIMU while keeping the visual covariance fixed (Table~\ref{tab:learned_imu_ablation}). Our covariance consistently outperforms AirIMU across all metrics, reducing the average gravity direction error from 0.611$\degree$ to 0.418$\degree$ (31.6\% $\downarrow$) and the body rotation error from 0.254$\degree$ to 0.215$\degree$ (15.4\% $\downarrow$). Since AirIMU's covariance misweights the inertial residuals during fusion, whereas our metrics-aware covariance reflects the reliability of each measurement and thus yields more accurate estimates.

\section{Conclusion} 
\label{sec:conclusion}

This paper presents MAC-I$^2$, which achieves robust VI fusion through learned metrics-aware covariances for both visual and inertial measurements, allowing vision and IMU to be weighted by their reliability rather than predefined uncertainties. As a showcase, we build a VI initialization and calibration system upon MAC-I$^2$, and demonstrate substantial gains in robustness and accuracy over state-of-the-art methods across challenging environments. In future work, we will extend MAC-I$^2$ to full visual-inertial odometry and SLAM, as well as broader multi-sensor fusion settings.






  




\bibliographystyle{IEEEtran}
\bibliography{./bibliography/references}

\clearpage
\newcommand{\rootloaded}{}
%
%

\ifdefined\rootloaded
  \DeclareRobustCommand{\maineqref}[1]{\eqref{#1}}   
\else
  \documentclass[letterpaper, 10 pt, conference]{ieeeconf}
  \IEEEoverridecommandlockouts
  \overrideIEEEmargins
  \usepackage{times}
  \usepackage{microtype}
  \usepackage{amsmath,amssymb,amsfonts,bm}
  \usepackage{gensymb}          
  \usepackage{nicefrac}
  \usepackage{graphicx}
  \usepackage{booktabs}
  \usepackage{multirow}
  \usepackage{float}
  \usepackage{caption}
  \usepackage[table]{xcolor}
  \usepackage[caption=false]{subfig}
  \usepackage{cite}
  \usepackage[bookmarks=true]{hyperref}
  \usepackage{cleveref}
  \usepackage{xr}
  \externaldocument[main-]{root}
  \DeclareRobustCommand{\maineqref}[1]{\eqref{main-#1}}
  \definecolor{ourgray}{gray}{0.9}
  \DeclareMathOperator{\Exp}{Exp}
  \DeclareMathOperator{\Log}{Log}
  \newcommand{\SO}{\mathrm{SO}}
  \newcommand{\SE}{\mathrm{SE}}
  \newcommand{\R}{\mathbb{R}}
  \newcommand{\bI}{\mathbf{I}}
  \newcommand{\bzero}{\mathbf{0}}
  \newcommand{\what}[1]{#1^{\wedge}}
  \newcommand{\Ad}{\mathrm{Ad}}
  \begin{document}
  \title{\LARGE \bf MAC-I$^2$: Learned Metrics-Aware Covariance for Robust\\
  Visual-Inertial Fusion in Initialization and Calibration\\ --- Appendix ---}
  \author{Paper-ID [XXXX]}
  \maketitle
  \IEEEpeerreviewmaketitle
\fi

\appendices

\section{Analytical Jacobians}
\label{app:analytical_jacobians}
\subsection{Notation}
\label{app:notation}
This appendix summarizes the analytical Jacobians used throughout our optimization. We linearize all residuals on \(\SO(3)\) and \(\SE(3)\) using a left-multiplicative perturbation model and report derivatives with respect to minimal tangent-space coordinates.

\paragraph{$\SE(3)$ exponential/logarithm}

For $\boldsymbol{\xi}=\begin{bmatrix}\boldsymbol{\rho}^\top & \boldsymbol{\phi}^\top\end{bmatrix}^\top\in\mathbb{R}^6$
we use $\Exp:\mathbb{R}^6\!\to\!\SE(3)$ and $\Log:\SE(3)\!\to\!\mathbb{R}^6$ with standard Lie definitions.
We denote the \emph{right/left Jacobians} on $\SE(3)$ by
$\mathbf{J}_r(\boldsymbol{\xi}),\mathbf{J}_l(\boldsymbol{\xi})\in\mathbb{R}^{6\times 6}$ and their inverses
$\mathbf{J}_r^{-1},\mathbf{J}_l^{-1}$.

\paragraph{Adjoint of $\SE(3)$}
For $\mathbf{T}=(\mathbf{p},\mathbf{R})\in\SE(3)$, the adjoint $\Ad_{\mathbf{T}}\in\mathbb{R}^{6\times 6}$ acting on
$\boldsymbol{\xi}=\begin{bmatrix}\boldsymbol{\rho}\\\boldsymbol{\phi}\end{bmatrix}$ (translation first) is
\begin{equation}
\begin{aligned}
\Ad_{\mathbf{T}} \triangleq {}&
\begin{bmatrix}
\mathbf{R} & [\mathbf{p}]_\times \mathbf{R}\\
\mathbf{0} & \mathbf{R}
\end{bmatrix},\\
\Ad_{\mathbf{T}}
\begin{bmatrix}\boldsymbol{\rho}\\\boldsymbol{\phi}\end{bmatrix}
={}&
\begin{bmatrix}
[\mathbf{p}]_\times \mathbf{R}\boldsymbol{\phi}+\mathbf{R}\boldsymbol{\rho}\\
\mathbf{R}\boldsymbol{\phi}
\end{bmatrix}.
\end{aligned}
\end{equation}

\subsection{Left perturbation model}
\label{app:left_perturb}

The optimization variables include (main paper Eq.~\maineqref{formula:state}):
\begin{equation}
\begin{aligned}
\mathbf{T}^{c_0}_{b_k}
&=
\begin{bmatrix}
\mathbf{R}^{c_0}_{b_k} & \mathbf{p}^{c_0}_{b_k}\\
\mathbf{0}^\top & 1
\end{bmatrix}\in\SE(3),\\
\mathbf{v}_{b_k}^{b_k} &\in\mathbb{R}^3,\quad
\mathbf{g}^{c_0}\in\mathbb{R}^3,\quad
\mathbf{b}_g,\mathbf{b}_a\in\mathbb{R}^3,\quad
\mathbf{T}^b_c\in\SE(3).
\end{aligned}
\end{equation}
We write $\mathbf{R}_k \triangleq \mathbf{R}^{c_0}_{b_k}$ and $\mathbf{p}_k \triangleq \mathbf{p}^{c_0}_{b_k}$.
The inverse rotation is $\mathbf{R}^{b_k}_{c_0}=\mathbf{R}_k^\top$.

\paragraph{Pose retraction (left)}
For $\delta\boldsymbol{\xi}_k=\begin{bmatrix}\delta\boldsymbol{\rho}_k\\\delta\boldsymbol{\phi}_k\end{bmatrix}\in\mathbb{R}^6$,
\begin{equation}
\mathbf{T}^{c_0}_{b_k} \leftarrow \Exp(\delta\boldsymbol{\xi}_k)\,\mathbf{T}^{c_0}_{b_k}.
\end{equation}
To first order,
\begin{equation}
\mathbf{R}_k \leftarrow \Exp(\delta\boldsymbol{\phi}_k)\mathbf{R}_k,
\qquad
\mathbf{p}_k \leftarrow \Exp(\delta\boldsymbol{\phi}_k)\mathbf{p}_k + \delta\boldsymbol{\rho}_k.
\label{eq:left_pose_update}
\end{equation}
Consequently,
\begin{equation}
\mathbf{R}_k^\top \leftarrow \mathbf{R}_k^\top \Exp(-\delta\boldsymbol{\phi}_k).
\label{eq:left_pose_update_invR}
\end{equation}

\paragraph{Euclidean variables}
We use additive perturbations:
\begin{equation}
\begin{aligned}
\mathbf{v}_{b_k}^{b_k} &\leftarrow \mathbf{v}_{b_k}^{b_k} + \delta\mathbf{v}_k,\\
\mathbf{g}^{c_0} &\leftarrow \mathbf{g}^{c_0} + \delta\mathbf{g},\\
\mathbf{b}_g &\leftarrow \mathbf{b}_g + \delta\mathbf{b}_g,\\
\mathbf{b}_a &\leftarrow \mathbf{b}_a + \delta\mathbf{b}_a.
\end{aligned}
\end{equation}

\subsection{First-order identities used in the derivations}
\label{app:idents}

\paragraph{Small-angle approximation}
\begin{equation}
\Exp(\delta\boldsymbol{\phi}) \approx \mathbf{I} + [\delta\boldsymbol{\phi}]_\times.
\end{equation}

\paragraph{Conjugation (SO(3))}
For any $\mathbf{R}\in\SO(3)$ and $\boldsymbol{\phi}\in\mathbb{R}^3$,
\begin{equation}
\Exp(\boldsymbol{\phi})\,\mathbf{R} = \mathbf{R}\,\Exp(\mathbf{R}^\top \boldsymbol{\phi}).
\label{eq:conj_so3}
\end{equation}

\paragraph{Log linearizations (SO(3))}
Let $\mathbf{E}\in\SO(3)$ and $\mathbf{r}=\Log(\mathbf{E}) \in \mathbb{R}^3$.
For small $\delta\in\mathbb{R}^3$:
\begin{align}
\Log(\Exp(\delta)\mathbf{E}) &\approx \mathbf{r} + \mathbf{J}_l^{-1}(\mathbf{r})\,\delta,
\label{eq:log_leftmult_so3}\\
\Log(\mathbf{E}\Exp(\delta)) &\approx \mathbf{r} + \mathbf{J}_r^{-1}(\mathbf{r})\,\delta.
\label{eq:log_rightmult_so3}
\end{align}

\paragraph{Log linearizations (SE(3))}
Let $\mathbf{T}\in\SE(3)$ and $\mathbf{r}=\Log(\mathbf{T})\in\mathbb{R}^6$.
For small $\delta\boldsymbol{\xi}\in\mathbb{R}^6$:
\begin{align}
\Log(\Exp(\delta\boldsymbol{\xi})\,\mathbf{T}) &\approx \mathbf{r} + \mathbf{J}_l^{-1}(\mathbf{r})\,\delta\boldsymbol{\xi},
\label{eq:log_leftmult_se3}\\
\Log(\mathbf{T}\Exp(\delta\boldsymbol{\xi})) &\approx \mathbf{r} + \mathbf{J}_r^{-1}(\mathbf{r})\,\delta\boldsymbol{\xi}.
\label{eq:log_rightmult_se3}
\end{align}

\subsection{Gyroscope bias and camera-IMU rotation estimation \texorpdfstring{(main paper Eq.~\maineqref{formula:gyro_extrinsic})}{}}
\label{app:gyro_extrinsic}

Let $\mathbf{q}_{c_k}^{c_{k+1}}$ be the visual relative rotation, and
$\boldsymbol{\gamma}^{b_k}_{b_{k+1}}\in\SO(3)$ the IMU preintegrated rotation.
The first-order bias correction is
\begin{equation}
\begin{aligned}
\hat{\boldsymbol{\gamma}}^{b_k}_{b_{k+1}}
&=
\boldsymbol{\gamma}^{b_k}_{b_{k+1}}\Exp(\boldsymbol{\phi}_g),\\
\boldsymbol{\phi}_g &\triangleq \mathbf{J}_{b_g}^\gamma\,\delta\mathbf{b}_g \in \mathbb{R}^3.
\end{aligned}
\end{equation}

Define
\begin{equation}
\mathbf{M}_k(\mathbf{q}^b_c) \triangleq
\mathbf{q}^b_c\,\mathbf{q}_{c_k}^{c_{k+1}}\,\mathbf{q}^c_b
\in \SO(3).
\end{equation}
The $\SO(3)$ residual is
\begin{equation}
\mathbf{r}^{\mathrm{gyro}}_k
\triangleq
\Log\!\Big(
\mathbf{M}_k(\mathbf{q}^b_c)\,\hat{\boldsymbol{\gamma}}^{b_k}_{b_{k+1}}
\Big)\in\mathbb{R}^3.
\end{equation}

Perturb $\delta\mathbf{b}_g\leftarrow \delta\mathbf{b}_g+\Delta\mathbf{b}_g$.
Then $\boldsymbol{\phi}_g\leftarrow \boldsymbol{\phi}_g+\mathbf{J}_{b_g}^\gamma\Delta\mathbf{b}_g$ and, using the standard first-order composition
$\Exp(\boldsymbol{\phi}+\delta\boldsymbol{\phi})\approx \Exp(\boldsymbol{\phi})\Exp(\mathbf{J}_r(\boldsymbol{\phi})\delta\boldsymbol{\phi})$,
\begin{equation}
\hat{\boldsymbol{\gamma}}'
\approx
\hat{\boldsymbol{\gamma}}\,
\Exp\!\Big(
\mathbf{J}_r(\boldsymbol{\phi}_g)\mathbf{J}_{b_g}^\gamma\Delta\mathbf{b}_g
\Big).
\end{equation}
Let $\mathbf{E} = \mathbf{M}_k(\mathbf{q}^b_c)\,\hat{\boldsymbol{\gamma}}$ and
$\mathbf{E}' = \mathbf{M}_k(\mathbf{q}^b_c)\,\hat{\boldsymbol{\gamma}}'$.
We have $\mathbf{E}'=\mathbf{E}\Exp(\delta)$ with
$\delta=\mathbf{J}_r(\boldsymbol{\phi}_g)\mathbf{J}_{b_g}^\gamma\Delta\mathbf{b}_g$, and by \eqref{eq:log_rightmult_so3}
\begin{equation}
\begin{aligned}
\Delta \mathbf{r}^{\mathrm{gyro}}_k \approx {}&
\mathbf{J}_r^{-1}(\mathbf{r}^{\mathrm{gyro}}_k)\,\delta\\
={}&
\mathbf{J}_r^{-1}(\mathbf{r}^{\mathrm{gyro}}_k)\,
\mathbf{J}_r(\boldsymbol{\phi}_g)\,
\mathbf{J}_{b_g}^\gamma\,\Delta\mathbf{b}_g.
\end{aligned}
\end{equation}
Hence
\begin{equation}
\boxed{
\begin{aligned}
\frac{\partial \mathbf{r}^{\mathrm{gyro}}_k}{\partial\,\delta\mathbf{b}_g}
&=
\mathbf{J}_r^{-1}(\mathbf{r}^{\mathrm{gyro}}_k)\,
\mathbf{J}_r(\boldsymbol{\phi}_g) \mathbf{J}_{b_g}^\gamma
\end{aligned}}
\end{equation}

Use a left perturbation $\mathbf{q}^b_c \leftarrow \Exp(\delta\boldsymbol{\theta})\mathbf{q}^b_c$.
Then $\mathbf{q}^c_b \leftarrow \mathbf{q}^c_b\Exp(-\delta\boldsymbol{\theta})$ and
\begin{align}
\mathbf{E}'
&=
\Exp(\delta\boldsymbol{\theta})\,\mathbf{M}_k\,\Exp(-\delta\boldsymbol{\theta})\,\hat{\boldsymbol{\gamma}}
\notag\\
&=
\Exp(\delta\boldsymbol{\theta})\,\mathbf{E}\,
\Big(\hat{\boldsymbol{\gamma}}^\top\Exp(-\delta\boldsymbol{\theta})\hat{\boldsymbol{\gamma}}\Big).
\label{eq:Eprime_theta}
\end{align}
Using $\hat{\boldsymbol{\gamma}}^\top\Exp(-\delta\boldsymbol{\theta})\hat{\boldsymbol{\gamma}}
=\Exp(-\hat{\boldsymbol{\gamma}}^\top\delta\boldsymbol{\theta})$ (from \eqref{eq:conj_so3}),
\begin{equation}
\mathbf{E}' =
\Exp(\delta\boldsymbol{\theta})\,\mathbf{E}\,\Exp(-\hat{\boldsymbol{\gamma}}^\top\delta\boldsymbol{\theta}).
\end{equation}
Applying \eqref{eq:log_leftmult_so3} and \eqref{eq:log_rightmult_so3} and dropping second-order terms yields
\begin{equation}
\begin{aligned}
\Delta \mathbf{r}^{\mathrm{gyro}}_k
\approx {}&
\mathbf{J}_l^{-1}(\mathbf{r}^{\mathrm{gyro}}_k)\,\delta\boldsymbol{\theta}-
\mathbf{J}_r^{-1}(\mathbf{r}^{\mathrm{gyro}}_k)\,\hat{\boldsymbol{\gamma}}^\top\delta\boldsymbol{\theta}.
\end{aligned}
\end{equation}
Hence
\begin{equation}
\boxed{
\begin{aligned}
\frac{\partial \mathbf{r}^{\mathrm{gyro}}_k}{\partial\,\delta\boldsymbol{\theta}}
&=
\mathbf{J}_l^{-1}(\mathbf{r}^{\mathrm{gyro}}_k)
-
\mathbf{J}_r^{-1}(\mathbf{r}^{\mathrm{gyro}}_k)\,\hat{\boldsymbol{\gamma}}^\top
\end{aligned}}
\end{equation}

\subsection{Visual pose-level residual \texorpdfstring{(main paper Eq.~\maineqref{formula:visual_res})}{}}
\label{app:visual_res}

Main paper Eq.~\maineqref{formula:visual_res} defines the pose-level visual residual (SE(3)):
\begin{equation}
\begin{aligned}
\mathbf{r}^{\mathrm{vis}}_k
\triangleq
\Log\!\Big(
&\mathbf{T}^{c_k}_{c_{k+1}}\;
\mathbf{T}^c_b\;
(\mathbf{T}^{c_0}_{b_{k+1}})^{-1}\mathbf{T}^{c_0}_{b_k}\;
\mathbf{T}^b_c
\Big)\in\mathbb{R}^6,
\end{aligned}
\end{equation}
where $\mathbf{T}^c_b=(\mathbf{T}^b_c)^{-1}$ and $\mathbf{T}^{c_k}_{c_{k+1}}$ is a measured visual relative motion.

Define the shorthand
\begin{equation}
\mathbf{A}_k \triangleq
\mathbf{T}^{c_k}_{c_{k+1}}\mathbf{T}^c_b(\mathbf{T}^{c_0}_{b_{k+1}})^{-1},
\end{equation}
\begin{equation}
\mathbf{B}_k \triangleq \mathbf{T}^{c_0}_{b_k}\mathbf{T}^b_c,
\end{equation}
\begin{equation}
\mathbf{T}_{\mathrm{err}} \triangleq \mathbf{A}_k\mathbf{B}_k,
\qquad
\mathbf{r}^{\mathrm{vis}}_k=\Log(\mathbf{T}_{\mathrm{err}}).
 \end{equation}

Perturb $\mathbf{T}^{c_0}_{b_k}\leftarrow \Exp(\delta\boldsymbol{\xi}_k)\mathbf{T}^{c_0}_{b_k}$.
Then
\begin{align}
\mathbf{T}'_{\mathrm{err}}
&=
\mathbf{A}_k\,\Exp(\delta\boldsymbol{\xi}_k)\,\mathbf{B}_k
\notag\\
&=
\big(\mathbf{A}_k\Exp(\delta\boldsymbol{\xi}_k)\mathbf{A}_k^{-1}\big)\,\mathbf{T}_{\mathrm{err}}
\notag\\
&=
\Exp(\Ad_{\mathbf{A}_k}\delta\boldsymbol{\xi}_k)\,\mathbf{T}_{\mathrm{err}}.
\end{align}
Using \eqref{eq:log_leftmult_se3} with $\mathbf{r}=\mathbf{r}^{\mathrm{vis}}_k$,
\begin{equation}
\Delta \mathbf{r}^{\mathrm{vis}}_k
\approx
\mathbf{J}_l^{-1}(\mathbf{r}^{\mathrm{vis}}_k)\,\Ad_{\mathbf{A}_k}\,\delta\boldsymbol{\xi}_k,
\end{equation}
hence
\begin{equation}
\boxed{
\begin{aligned}
\frac{\partial \mathbf{r}^{\mathrm{vis}}_k}{\partial\,\delta\boldsymbol{\xi}_k}
&=
\mathbf{J}_l^{-1}(\mathbf{r}^{\mathrm{vis}}_k)\,\Ad_{\mathbf{A}_k}
\end{aligned}}
\end{equation}

Perturb $\mathbf{T}^{c_0}_{b_{k+1}}\leftarrow \Exp(\delta\boldsymbol{\xi}_{k+1})\mathbf{T}^{c_0}_{b_{k+1}}$.
Then
$(\mathbf{T}^{c_0}_{b_{k+1}})^{-1}\leftarrow (\mathbf{T}^{c_0}_{b_{k+1}})^{-1}\Exp(-\delta\boldsymbol{\xi}_{k+1})$ and
\begin{align}
\mathbf{T}'_{\mathrm{err}}
&=
\mathbf{A}_k\,\Exp(-\delta\boldsymbol{\xi}_{k+1})\,\mathbf{B}_k
\notag\\
&=
\Exp(\Ad_{\mathbf{A}_k}(-\delta\boldsymbol{\xi}_{k+1}))\,\mathbf{T}_{\mathrm{err}}.
\end{align}
Therefore
\begin{equation}
\boxed{
\begin{aligned}
\frac{\partial \mathbf{r}^{\mathrm{vis}}_k}{\partial\,\delta\boldsymbol{\xi}_{k+1}}
&=
-\mathbf{J}_l^{-1}(\mathbf{r}^{\mathrm{vis}}_k)\,\Ad_{\mathbf{A}_k}
\end{aligned}}
\end{equation}

Use left perturbation $\mathbf{T}^b_c \leftarrow \Exp(\delta\boldsymbol{\xi}_{bc})\mathbf{T}^b_c$.
Then $\mathbf{T}^c_b\leftarrow \mathbf{T}^c_b\Exp(-\delta\boldsymbol{\xi}_{bc})$ and
\begin{align}
\mathbf{T}'_{\mathrm{err}}
&=
\underbrace{\mathbf{T}^{c_k}_{c_{k+1}}\mathbf{T}^c_b}_{\triangleq\,\mathbf{S}_k}
\Exp(-\delta\boldsymbol{\xi}_{bc})
(\mathbf{T}^{c_0}_{b_{k+1}})^{-1}\mathbf{T}^{c_0}_{b_k}
\Exp(\delta\boldsymbol{\xi}_{bc})\mathbf{T}^b_c.
\end{align}
Let $\mathbf{C}\triangleq \mathbf{T}_b^c$ and note that
$\mathbf{T}_{\mathrm{err}}=\mathbf{S}_k(\mathbf{T}^{c_0}_{b_{k+1}})^{-1}\mathbf{T}^{c_0}_{b_k}\mathbf{C}$.
We have
\begin{align}
\mathbf{T}'_{\mathrm{err}}
&=
\big(\mathbf{S}_k\Exp(-\delta\boldsymbol{\xi}_{bc})\mathbf{S}_k^{-1}\big)\;
\mathbf{T}_{\mathrm{err}}\;
\big(\mathbf{C}^{-1}\Exp(\delta\boldsymbol{\xi}_{bc})\mathbf{C}\big)
\notag\\
&=
\Exp(\Ad_{\mathbf{S}_k}(-\delta\boldsymbol{\xi}_{bc}))\;
\mathbf{T}_{\mathrm{err}}\;
\Exp(\Ad_{\mathbf{C}^{-1}}\delta\boldsymbol{\xi}_{bc}).
\end{align}
Using \eqref{eq:log_leftmult_se3} and \eqref{eq:log_rightmult_se3}:
\begin{equation}
\begin{aligned}
\Delta \mathbf{r}^{\mathrm{vis}}_k
\approx {}&
\mathbf{J}_r^{-1}(\mathbf{r}^{\mathrm{vis}}_k)\,\Ad_{\mathbf{C}^{-1}}\delta\boldsymbol{\xi}_{bc}\\
&+
\mathbf{J}_l^{-1}(\mathbf{r}^{\mathrm{vis}}_k)\,\Ad_{\mathbf{S}_k}\delta(-\delta\boldsymbol{\xi}_{bc}),
\end{aligned}
\end{equation}
hence
\begin{equation}
\boxed{
\begin{aligned}
\frac{\partial \mathbf{r}^{\mathrm{vis}}_k}{\partial\,\delta\boldsymbol{\xi}_{bc}}
&=
\mathbf{J}_r^{-1}(\mathbf{r}^{\mathrm{vis}}_k)\,\Ad_{\mathbf{C}^{-1}}\\
&\quad-
\mathbf{J}_l^{-1}(\mathbf{r}^{\mathrm{vis}}_k)\,\Ad_{\mathbf{S}_k}
\end{aligned}}
\end{equation}

\textbf{Other state Jacobians:}
$\mathbf{r}^{\mathrm{vis}}_k$ does not depend on $\mathbf{v}_{b_k}^{b_k}$, $\mathbf{g}^{c_0}$, $\mathbf{b}_g$, or $\mathbf{b}_a$:
their Jacobians are $\mathbf{0}$.

\subsection{IMU residuals \texorpdfstring{(main paper Eqs.~\maineqref{formula:inte_res_r}--\maineqref{formula:inte_res_p})}{}}
\label{app:imu_residuals}

Let $\Delta t \triangleq \Delta t_{kk+1}$.
Recall $\mathbf{R}_k=\mathbf{R}^{c_0}_{b_k}$, $\mathbf{R}^{b_k}_{c_0}=\mathbf{R}_k^\top$.
We denote $\mathbf{v}_k\triangleq \mathbf{v}_{b_k}^{b_k}$ and $\mathbf{g}\triangleq \mathbf{g}^{c_0}$ for brevity.
In the main paper, $\tilde{\boldsymbol{\gamma}}^{b_k}_{b_{k+1}}, \tilde{\boldsymbol{\beta}}^{b_k}_{b_{k+1}}, \tilde{\boldsymbol{\alpha}}^{b_k}_{b_{k+1}}$ denote the preintegration terms of whichever workflow is used; below we spell out the two cases separately, since the no-learned-IMU case additionally depends on the bias increments.

\subsubsection{Rotation residual \texorpdfstring{$\mathbf{r}_{\gamma_k}$ (main paper Eq.~\maineqref{formula:inte_res_r})}{r\_gamma\_k}}
\label{app:r_gamma}

\paragraph{Learned-IMU case}
Define
\begin{equation}
\begin{aligned}
\mathbf{E}_\gamma
&\triangleq
\mathbf{R}^{b_{k+1}}_{c_0}\mathbf{R}^{c_0}_{b_k}\,\tilde{\boldsymbol{\gamma}}^{b_k}_{b_{k+1}}
=
\mathbf{R}_{k+1}^\top \mathbf{R}_k \,\tilde{\boldsymbol{\gamma}}^{b_k}_{b_{k+1}},\\
\mathbf{r}_{\gamma k} &\triangleq \Log(\mathbf{E}_\gamma)\in\mathbb{R}^3.
\end{aligned}
\end{equation}

Apply left perturbations $\mathbf{R}_k\leftarrow \Exp(\delta\boldsymbol{\phi}_k)\mathbf{R}_k$ and
$\mathbf{R}_{k+1}\leftarrow \Exp(\delta\boldsymbol{\phi}_{k+1})\mathbf{R}_{k+1}$.
Using \eqref{eq:left_pose_update_invR},
\begin{align}
\mathbf{E}'_\gamma
&=
(\mathbf{R}_{k+1}')^\top \mathbf{R}'_k \tilde{\boldsymbol{\gamma}}
\notag\\
&=
\mathbf{R}_{k+1}^\top \Exp(-\delta\boldsymbol{\phi}_{k+1})
\Exp(\delta\boldsymbol{\phi}_k)\mathbf{R}_k \tilde{\boldsymbol{\gamma}}
\notag\\
&\approx
\mathbf{R}_{k+1}^\top \Exp(\delta\boldsymbol{\phi}_k-\delta\boldsymbol{\phi}_{k+1})\mathbf{R}_k \tilde{\boldsymbol{\gamma}}
\notag\\
&\overset{\eqref{eq:conj_so3}}{=}
\Exp\!\Big(\mathbf{R}_{k+1}^\top(\delta\boldsymbol{\phi}_k-\delta\boldsymbol{\phi}_{k+1})\Big)\;
\mathbf{R}_{k+1}^\top\mathbf{R}_k \tilde{\boldsymbol{\gamma}}
\notag\\
&=
\Exp\!\Big(\mathbf{R}^{b_{k+1}}_{c_0}(\delta\boldsymbol{\phi}_k-\delta\boldsymbol{\phi}_{k+1})\Big)\;\mathbf{E}_\gamma.
\end{align}
Then by \eqref{eq:log_leftmult_so3},
\begin{equation}
\Delta\mathbf{r}_{\gamma_k}
\approx
\mathbf{J}_l^{-1}(\mathbf{r}_{\gamma k})\,\mathbf{R}^{b_{k+1}}_{c_0}(\delta\boldsymbol{\phi}_k-\delta\boldsymbol{\phi}_{k+1}).
\end{equation}
Therefore
\begin{equation}
\boxed{
\begin{aligned}
\frac{\partial \mathbf{r}_{\gamma_k}}{\partial\,\delta\boldsymbol{\phi}_k}
&=
\mathbf{J}_l^{-1}(\mathbf{r}_{\gamma_k})\,\mathbf{R}^{b_{k+1}}_{c_0},\\
\frac{\partial \mathbf{r}_{\gamma_k}}{\partial\,\delta\boldsymbol{\phi}_{k+1}}
&=
-\mathbf{J}_l^{-1}(\mathbf{r}_{\gamma_k})\,\mathbf{R}^{b_{k+1}}_{c_0}
\end{aligned}}
\end{equation}
and
\begin{equation}
\boxed{
\begin{aligned}
\frac{\partial \mathbf{r}_{\gamma_k}}{\partial\,\delta\boldsymbol{\rho}_k}
&=\mathbf{0},\\
\frac{\partial \mathbf{r}_{\gamma_k}}{\partial\,\delta\boldsymbol{\rho}_{k+1}}
&=\mathbf{0}.
\end{aligned}}
\end{equation}

\paragraph{No-learned-IMU case}
Now
\begin{equation}
\begin{aligned}
\mathbf{r}_{\gamma_k}
&=
\Log\!\Big(
\mathbf{R}^{b_{k+1}}_{c_0}\mathbf{R}^{c_0}_{b_k}\,
\boldsymbol{\gamma}^{b_k}_{b_{k+1}}\Exp(\boldsymbol{\phi}_g)
\Big),\\
\boldsymbol{\phi}_g &\triangleq \mathbf{J}_{b_g}^\gamma\,\delta\mathbf{b}_g.
\end{aligned}
\end{equation}
The pose Jacobians w.r.t.\ $\delta\boldsymbol{\phi}_k,\delta\boldsymbol{\phi}_{k+1}$ are identical to the learned case above
(replace $\tilde{\boldsymbol{\gamma}}$ by $\boldsymbol{\gamma}\Exp(\boldsymbol{\phi}_g)$).

For the gyro bias, perturb $\delta\mathbf{b}_g\leftarrow \delta\mathbf{b}_g+\Delta\mathbf{b}_g$.
Then $\Exp(\boldsymbol{\phi}_g)$ right-multiplies by $\Exp(\mathbf{J}_r(\boldsymbol{\phi}_g)\mathbf{J}_{b_g}^\gamma\Delta\mathbf{b}_g)$,
hence by \eqref{eq:log_rightmult_so3}
\begin{equation}
\boxed{
\begin{aligned}
\frac{\partial \mathbf{r}_{\gamma_k}}{\partial\,\delta\mathbf{b}_g}
&=
\mathbf{J}_r^{-1}(\mathbf{r}_{\gamma_k})\,
\mathbf{J}_r(\boldsymbol{\phi}_g)\,\mathbf{J}_{b_g}^\gamma
\end{aligned}}
\end{equation}

\subsubsection{Velocity residual \texorpdfstring{$\mathbf{r}_{\beta_k}$ (main paper Eq.~\maineqref{formula:inte_res_v})}{r\_beta\_k}}
\label{app:r_beta}

\paragraph{Learned-IMU case}
Write
\begin{align}
\mathbf{r}_{\beta_k}
={}&
\mathbf{R}^{b_k}_{c_0}\mathbf{R}^{c_0}_{b_{k+1}}\mathbf{v}_{k+1}
-\mathbf{v}_k
-\mathbf{R}^{b_k}_{c_0}\mathbf{g}\Delta t
-\tilde{\boldsymbol{\beta}}^{b_k}_{b_{k+1}}
\notag\\
={}&
\mathbf{R}_k^\top(\mathbf{R}_{k+1}\mathbf{v}_{k+1}-\mathbf{g}\Delta t)
-\mathbf{v}_k-\tilde{\boldsymbol{\beta}}.
\end{align}
Define $\mathbf{a}\triangleq \mathbf{R}_{k+1}\mathbf{v}_{k+1}\in\mathbb{R}^3$ (velocity expressed in $c_0$).
Using \eqref{eq:left_pose_update_invR}, $\mathbf{R}_k^\top\leftarrow \mathbf{R}_k^\top\Exp(-\delta\boldsymbol{\phi}_k)$ implies
\begin{equation}
\begin{aligned}
\mathbf{R}_k^\top \Exp(-\delta\boldsymbol{\phi}_k)\mathbf{y}
&\approx
\mathbf{R}_k^\top(\mathbf{y}+[\mathbf{y}]_\times \delta\boldsymbol{\phi}_k),\\
\Rightarrow\quad
\delta(\mathbf{R}_k^\top\mathbf{y})
&=
\mathbf{R}_k^\top[\mathbf{y}]_\times\delta\boldsymbol{\phi}_k.
\end{aligned}
\end{equation}
Also, $\mathbf{a}'=\Exp(\delta\boldsymbol{\phi}_{k+1})\mathbf{a}\approx \mathbf{a}-[\mathbf{a}]_\times\delta\boldsymbol{\phi}_{k+1}$.

Collecting first-order terms,
\begin{equation}
\boxed{
\begin{aligned}
\frac{\partial \mathbf{r}_{\beta_k}}{\partial\,\delta\boldsymbol{\phi}_k}
&=
\mathbf{R}_k^\top[\mathbf{a}]_\times
-
\Delta t\,\mathbf{R}_k^\top[\mathbf{g}]_\times,\\
\frac{\partial \mathbf{r}_{\beta_k}}{\partial\,\delta\boldsymbol{\phi}_{k+1}}
&=
-\mathbf{R}_k^\top[\mathbf{a}]_\times
\end{aligned}}
\end{equation}
and
\begin{equation}
\boxed{
\begin{aligned}
\frac{\partial \mathbf{r}_{\beta_k}}{\partial\,\delta\boldsymbol{\rho}_k}
&=\mathbf{0},\\
\frac{\partial \mathbf{r}_{\beta_k}}{\partial\,\delta\boldsymbol{\rho}_{k+1}}
&=\mathbf{0}.
\end{aligned}}
\end{equation}
Euclidean Jacobians:
\begin{equation}
\boxed{
\begin{aligned}
\frac{\partial \mathbf{r}_{\beta_k}}{\partial\,\mathbf{v}_k}
&=-\mathbf{I},\\
\frac{\partial \mathbf{r}_{\beta_k}}{\partial\,\mathbf{v}_{k+1}}
&=\mathbf{R}_k^\top\mathbf{R}_{k+1},\\
\frac{\partial \mathbf{r}_{\beta_k}}{\partial\,\mathbf{g}}
&=-\Delta t\,\mathbf{R}_k^\top.
\end{aligned}}
\end{equation}

\paragraph{No-learned-IMU case}
\begin{equation}
\begin{aligned}
\mathbf{r}_{\beta_k}
&=
\mathbf{R}_k^\top\mathbf{R}_{k+1}\mathbf{v}_{k+1}
-\mathbf{v}_k
-\mathbf{R}_k^\top\mathbf{g}\Delta t\\
&\quad-
\Big(\boldsymbol{\beta}+\mathbf{J}_{b_g}^\beta\delta\mathbf{b}_g+\mathbf{J}_{b_a}^\beta\delta\mathbf{b}_a\Big).
\end{aligned}
\end{equation}
All Jacobians above remain unchanged, and additionally
\begin{equation}
\boxed{
\begin{aligned}
\frac{\partial \mathbf{r}_{\beta_k}}{\partial\,\delta\mathbf{b}_g}
&=-\mathbf{J}_{b_g}^\beta,\\
\frac{\partial \mathbf{r}_{\beta_k}}{\partial\,\delta\mathbf{b}_a}
&=-\mathbf{J}_{b_a}^\beta.
\end{aligned}}
\end{equation}

\subsubsection{Position residual \texorpdfstring{$\mathbf{r}_{\alpha_k}$ (main paper Eq.~\maineqref{formula:inte_res_p})}{r\_alpha\_k}}
\label{app:r_alpha}

\paragraph{Learned-IMU case}
\begin{equation}
\begin{aligned}
\mathbf{r}_{\alpha_k}
&=
\mathbf{R}^{b_k}_{c_0}\mathbf{p}_{k+1}
-\mathbf{R}^{b_k}_{c_0}\mathbf{p}_k
-\mathbf{v}_k\Delta t
-\tfrac12\,\mathbf{R}^{b_k}_{c_0}\mathbf{g}\Delta t^2
-\tilde{\boldsymbol{\alpha}}^{b_k}_{b_{k+1}}\\
&=
\mathbf{R}_k^\top\mathbf{p}_{k+1}
-\mathbf{R}_k^\top\mathbf{p}_k
-\mathbf{v}_k\Delta t
-\tfrac12\mathbf{R}_k^\top\mathbf{g}\Delta t^2
-\tilde{\boldsymbol{\alpha}}.
\end{aligned}
\end{equation}

The \emph{left} pose update \eqref{eq:left_pose_update} rotates $\mathbf{p}_{k+1}$:
\begin{equation}
\begin{aligned}
\mathbf{p}_{k+1}
&\leftarrow
\Exp(\delta\boldsymbol{\phi}_{k+1})\mathbf{p}_{k+1}+\delta\boldsymbol{\rho}_{k+1}\\
&\approx
\mathbf{p}_{k+1}-[\mathbf{p}_{k+1}]_\times\delta\boldsymbol{\phi}_{k+1}+\delta\boldsymbol{\rho}_{k+1}.
\end{aligned}
\end{equation}

\paragraph{Jacobian w.r.t.\ pose at $k$}
Using \eqref{eq:left_pose_update} and \eqref{eq:left_pose_update_invR}, one can check that the first-order $\delta\boldsymbol{\phi}_k$ terms in
$\mathbf{R}_k^\top\mathbf{p}_k$ cancel (because both $\mathbf{R}_k$ and $\mathbf{p}_k$ are rotated by the same left perturbation),
while $\mathbf{R}_k^\top\mathbf{p}_{k+1}$ and $\mathbf{R}_k^\top\mathbf{g}$ retain $\delta\boldsymbol{\phi}_k$ sensitivity. The result is
\begin{equation}
\boxed{
\begin{aligned}
\frac{\partial \mathbf{r}_{\alpha_k}}{\partial\,\delta\boldsymbol{\phi}_k}
&=
\mathbf{R}_k^\top\Big([\mathbf{p}_{k+1}]_\times-\tfrac12\Delta t^2[\mathbf{g}]_\times\Big),\\
\frac{\partial \mathbf{r}_{\alpha_k}}{\partial\,\delta\boldsymbol{\rho}_k}
&=
-\mathbf{R}_k^\top.
\end{aligned}}
\end{equation}

\paragraph{Jacobian w.r.t.\ pose at $k{+}1$}
Only $\mathbf{p}_{k+1}$ contributes:
\begin{equation}
\boxed{
\begin{aligned}
\frac{\partial \mathbf{r}_{\alpha_k}}{\partial\,\delta\boldsymbol{\phi}_{k+1}}
&=
-\mathbf{R}_k^\top[\mathbf{p}_{k+1}]_\times,\\
\frac{\partial \mathbf{r}_{\alpha_k}}{\partial\,\delta\boldsymbol{\rho}_{k+1}}
&=
\mathbf{R}_k^\top.
\end{aligned}}
\end{equation}

\paragraph{Euclidean Jacobians.}
\begin{equation}
\boxed{
\begin{aligned}
\frac{\partial \mathbf{r}_{\alpha_k}}{\partial\,\mathbf{v}_k}
&=-\Delta t\,\mathbf{I},\\
\frac{\partial \mathbf{r}_{\alpha_k}}{\partial\,\mathbf{g}}
&=-\tfrac12\Delta t^2\,\mathbf{R}_k^\top.
\end{aligned}}
\end{equation}

\paragraph{No-learned-IMU case.}
\begin{equation}
\begin{aligned}
\mathbf{r}_{\alpha k}
&=
\mathbf{R}_k^\top\mathbf{p}_{k+1}
-\mathbf{R}_k^\top\mathbf{p}_k
-\mathbf{v}_k\Delta t
-\tfrac12\mathbf{R}_k^\top\mathbf{g}\Delta t^2\\
&\quad-
\Big(\boldsymbol{\alpha}+\mathbf{J}_{b_g}^\alpha\delta\mathbf{b}_g+\mathbf{J}_{b_a}^\alpha\delta\mathbf{b}_a\Big).
\end{aligned}
\end{equation}
All Jacobians above remain unchanged, and additionally
\begin{equation}
\boxed{
\begin{aligned}
\frac{\partial \mathbf{r}_{\alpha k}}{\partial\,\delta\mathbf{b}_g}
&=-\mathbf{J}_{b_g}^\alpha,\\
\frac{\partial \mathbf{r}_{\alpha k}}{\partial\,\delta\mathbf{b}_a}
&=-\mathbf{J}_{b_a}^\alpha.
\end{aligned}}
\end{equation}

\section{Visual Residual Covariance}
\label{app:visual_res_cov}

The visual residual of main paper Eq.~\maineqref{formula:visual_res} is built from the relative
pose $\mathbf{T}^{c_k}_{c_{k+1}}$ estimated by MAC-VO, so its covariance is
obtained in two steps: we first propagate the learned $3$D keypoint
uncertainties through the pose graph optimization (PGO) to a covariance on the
estimated relative pose (\cref{app:pose_cov_derivation}), and then propagate
that pose covariance through the $\Log$ map that defines the residual
(\cref{app:res_cov_propagation}). Throughout we keep the left-perturbation
convention and the translation-first ordering
$\delta\boldsymbol{\xi}=\begin{bmatrix}\delta\boldsymbol{\rho}^\top&\delta\boldsymbol{\phi}^\top\end{bmatrix}^\top$
of \cref{app:left_perturb}.

\subsection{Covariance of the estimated relative visual pose}
\label{app:pose_cov_derivation}

MAC-VO estimates the relative pose between consecutive frames by solving the
PGO of main paper Eq.~\maineqref{formula:macvo}. Let
$\hat{\mathbf{T}}\triangleq \mathbf{T}^{c_{k}}_{c_{k+1}}{}^\star$ be its
minimizer, with rotation part $\hat{\mathbf{R}}$, and parametrize a
neighbourhood of $\hat{\mathbf{T}}$ by the left perturbation
\begin{equation}
\mathbf{T}(\delta\boldsymbol{\xi}) = \Exp(\delta\boldsymbol{\xi})\,\hat{\mathbf{T}},
\qquad \delta\boldsymbol{\xi}\in\mathbb{R}^6 .
\label{eq:vis_pose_perturb}
\end{equation}
The per-keypoint residual of the PGO is
\begin{equation}
\mathbf{r}_{i,k}(\delta\boldsymbol{\xi})
=
\mathbf{p}^{c}_{k,i}
-
\Exp(\delta\boldsymbol{\xi})\,\hat{\mathbf{T}}\,\mathbf{p}^{c}_{k+1,i}
\in\mathbb{R}^3 .
\label{eq:vis_pgo_residual}
\end{equation}

\paragraph{Residual weight}
MAC-VO predicts a $3$D covariance $\bm{\Sigma}^p_{k,i}$ for every keypoint
$\mathbf{p}^{c}_{k,i}$. The covariance of
\eqref{eq:vis_pgo_residual} at $\delta\boldsymbol{\xi}=\mathbf{0}$ is, to first
order,
\begin{equation}
\bm{\Sigma}_i
=
\bm{\Sigma}^p_{k,i}
+
\hat{\mathbf{R}}\,\bm{\Sigma}^p_{k+1,i}\,\hat{\mathbf{R}}^\top ,
\label{eq:vis_match_cov}
\end{equation}

\paragraph{Linearization at the optimum}
Using $\Exp(\delta\boldsymbol{\xi})\mathbf{q}\approx
\mathbf{q}+\delta\boldsymbol{\rho}+[\delta\boldsymbol{\phi}]_\times\mathbf{q}
=\mathbf{q}+\delta\boldsymbol{\rho}-[\mathbf{q}]_\times\delta\boldsymbol{\phi}$
with $\mathbf{q}\triangleq \hat{\mathbf{T}}\mathbf{p}^{c}_{k+1,i}$, we obtain
\begin{equation}
\begin{aligned}
\mathbf{r}_{i,k}(\delta\boldsymbol{\xi})
&\approx
\mathbf{r}_{i,k}(\mathbf{0})
-\delta\boldsymbol{\rho}
+[\mathbf{q}]_\times\delta\boldsymbol{\phi}\\
&=
\mathbf{r}_{i,k}(\mathbf{0})+\mathbf{J}_{i,k}\,\delta\boldsymbol{\xi},
\end{aligned}
\end{equation}
so that
\begin{equation}
\boxed{
\mathbf{J}_{i,k}
=
\frac{\partial \mathbf{r}_{i,k}(\delta\boldsymbol{\xi})}{\partial\,\delta\boldsymbol{\xi}}
\bigg|_{\delta\boldsymbol{\xi}=\mathbf{0}}
=
\begin{bmatrix}
-\mathbf{I}_{3\times3} & [\hat{\mathbf{T}}\mathbf{p}^{c}_{k+1,i}]_\times
\end{bmatrix}}
\label{eq:vis_pgo_jac}
\end{equation}
which is the Jacobian reported in main paper Eq.~\maineqref{pose_cov}.

\paragraph{Laplace approximation}
The negative log-likelihood of the PGO is
\begin{equation}
\mathcal{L}(\delta\boldsymbol{\xi})
=
\tfrac12\sum_i
\mathbf{r}_{i,k}(\delta\boldsymbol{\xi})^\top
\bm{\Sigma}_i^{-1}
\mathbf{r}_{i,k}(\delta\boldsymbol{\xi})
+\text{const}.
\end{equation}
Substituting the linearization \eqref{eq:vis_pgo_jac} gives the quadratic model
\begin{equation}
\begin{aligned}
\mathcal{L}(\delta\boldsymbol{\xi})
&\approx
\mathcal{L}(\mathbf{0})
+\mathbf{g}_k^\top\delta\boldsymbol{\xi}
+\tfrac12\,\delta\boldsymbol{\xi}^\top\mathbf{H}_k\,\delta\boldsymbol{\xi},\\
\mathbf{g}_k &= \sum_i \mathbf{J}_{i,k}^\top\bm{\Sigma}_i^{-1}\mathbf{r}_{i,k}(\mathbf{0}),\quad
\mathbf{H}_k \approx \sum_i \mathbf{J}_{i,k}^\top\bm{\Sigma}_i^{-1}\mathbf{J}_{i,k},
\end{aligned}
\label{eq:vis_quadratic_model}
\end{equation}
where $\mathbf{H}_k$ is the Gauss-Newton (Fisher information) approximation of
the Hessian, i.e.\ the term involving
$\partial^2\mathbf{r}_{i,k}/\partial\delta\boldsymbol{\xi}^2$ is dropped, which
is standard and vanishes at a zero-residual optimum. Because
$\hat{\mathbf{T}}$ is a stationary point, $\mathbf{g}_k\approx\mathbf{0}$, and
$\mathcal{L}$ reduces to a centered quadratic form. Exponentiating it yields the
Laplace approximation of the posterior,
$\delta\boldsymbol{\xi}\sim\mathcal{N}(\mathbf{0},\bm{\Sigma}_{\mathbf{T}^{c_{k}}_{c_{k+1}}})$,
whose covariance is the inverse of that quadratic form:
\begin{equation}
\boxed{\;
\bm{\Sigma}_{\mathbf{T}^{c_{k}}_{c_{k+1}}}
=
\mathbf{H}_k^{-1}\;}
\label{eq:vis_pose_cov}
\end{equation}
which is main paper Eq.~\maineqref{pose_cov}. Note that
$\bm{\Sigma}_{\mathbf{T}^{c_{k}}_{c_{k+1}}}$ lives in the tangent space at
$\hat{\mathbf{T}}$ under the \emph{left} perturbation
\eqref{eq:vis_pose_perturb}.

\subsection{Propagation to the pose-level residual}
\label{app:res_cov_propagation}

The estimated relative pose enters the pose-level residual of main paper
Eq.~\maineqref{formula:visual_res} as a \emph{measurement}. Writing the underlying
true relative motion as a left perturbation of the estimate,
\begin{equation}
\mathbf{T}^{c_k}_{c_{k+1}}
=
\Exp(\delta\boldsymbol{\xi})\,\hat{\mathbf{T}}^{c_k}_{c_{k+1}},
\qquad
\delta\boldsymbol{\xi}\sim\mathcal{N}(\mathbf{0},\bm{\Sigma}_{\mathbf{T}^{c_{k}}_{c_{k+1}}}),
\end{equation}
and reusing $\mathbf{T}_{\mathrm{err}}$ of \cref{app:visual_res} evaluated at
the measured pose, the perturbed residual is
\begin{equation}
\begin{aligned}
\mathbf{r}^{\mathrm{vis}}_k(\delta\boldsymbol{\xi})
&=
\Log\!\Big(
\Exp(\delta\boldsymbol{\xi})\,
\hat{\mathbf{T}}^{c_k}_{c_{k+1}}\mathbf{T}^c_b
(\mathbf{T}^{c_0}_{b_{k+1}})^{-1}\mathbf{T}^{c_0}_{b_k}\mathbf{T}^b_c
\Big)\\
&=
\Log\!\big(\Exp(\delta\boldsymbol{\xi})\,\mathbf{T}_{\mathrm{err}}\big).
\end{aligned}
\end{equation}
The measured pose is the \emph{leftmost} factor of $\mathbf{T}_{\mathrm{err}}$,
so its perturbation needs no adjoint transport, in contrast with the state
Jacobians of \cref{app:visual_res} where the perturbed factor sits in the middle
and $\Ad_{\mathbf{A}_k}$ appears. Applying \eqref{eq:log_leftmult_se3} directly,
\begin{equation}
\mathbf{r}^{\mathrm{vis}}_k(\delta\boldsymbol{\xi})
\approx
\mathbf{r}^{\mathrm{vis}}_k
+
\mathbf{J}_l^{-1}(\mathbf{r}^{\mathrm{vis}}_k)\,\delta\boldsymbol{\xi},
\end{equation}
hence
\begin{equation}
\boxed{
\frac{\partial \mathbf{r}^{\mathrm{vis}}_k}{\partial\,\delta\boldsymbol{\xi}}
=
\mathbf{J}_l^{-1}(\mathbf{r}^{\mathrm{vis}}_k)}
\label{eq:vis_res_meas_jac}
\end{equation}
and, by linear propagation of the Gaussian
$\delta\boldsymbol{\xi}$ through \eqref{eq:vis_res_meas_jac},
\begin{equation}
\boxed{
\bm{\Sigma}^{\text{visual}}_k
=
\mathbf{J}_l^{-1}(\mathbf{r}^{\mathrm{vis}}_k)\,
\bm{\Sigma}_{\mathbf{T}^{c_{k}}_{c_{k+1}}}\,
\mathbf{J}_l^{-\top}(\mathbf{r}^{\mathrm{vis}}_k)}
\label{eq:vis_res_cov}
\end{equation}
which is main paper Eq.~\maineqref{formula:visual_res_cov}.

\section{IMU Residual Covariance}
The covariance of the IMU residual can be derived from the IMU preintegration covariance:
\begin{equation}
\begin{aligned}
\Sigma^r_{\gamma_k} &= J_r^{-1}(\mathbf{r}_{\gamma_k}) \Sigma^{\Delta R}_k (J_r^{-1}(\mathbf{r}_{\gamma_k}))^T, \\
\Sigma^v_{\beta_k} &= \Sigma^{\Delta v}_k, \\
\Sigma^p_{\alpha_k} &= \Sigma^{\Delta p}_k,
\end{aligned}
\end{equation}

\section{Learned IMU Model}

\begin{figure}[htbp]
\centering
\includegraphics[width=\columnwidth]{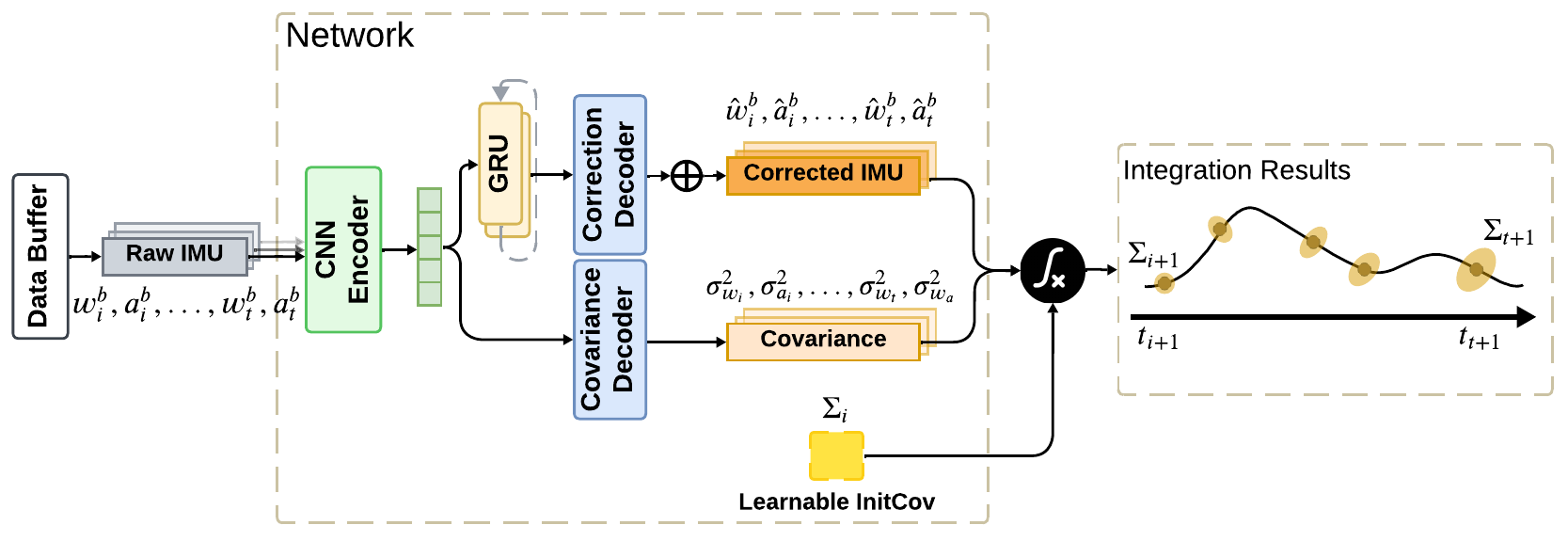}
\caption{Network architecture of the learned imu model developed in this paper.}
\label{fig:airimu_net}
\vspace{-5pt}
\end{figure}

\noindent \textbf{Network Architecture} The network architecture of the learned imu model is illustrated in Fig.~\ref{fig:airimu_net}. Following AirIMU, we train a shared CNN encoder network to leverage the hidden relationship between IMU correction and uncertainty, while we do not use GRU-processed data to predict the uncertainty. In addition, a learnable init covariance is introduced.

\noindent \textbf{IMU Preintegration Covariance Propagation} Our model jointly supervises the integrated IMU state and the propagated covariance to train the model. The preintegration covariance propagation is as follows:
\begin{equation}
\label{formula:cov_prop}
\begin{aligned}
\bm{\Sigma}_{ik+1} &= \mathbf{A}\bm{\Sigma}_{ik}\mathbf{A}^T + \mathbf{B}\text{diag}(\bm{\Sigma}^{gyro}, \bm{\Sigma}^{acc})\mathbf{B}^T \\
&= \mathbf{A}\bm{\Sigma}_{ik}\mathbf{A}^T + \mathbf{B}_g\bm{\Sigma}^{gyro}\mathbf{B}_g^T + \mathbf{B}_a\bm{\Sigma}^{acc}\mathbf{B}_a^T,
\end{aligned}
\end{equation}
where $\bm{\Sigma}^{gyro}$ and $\bm{\Sigma}^{acc}$ are the measurement covariance, which is obtained with the uncertainty prediction of the network. The propagation starts from $\bm{\Sigma}_{ii} = \bm{\Sigma}_{0}$, where $\bm{\Sigma}_{0}$ is the learned initial covariance. $\mathbf{A}$ and $\mathbf{B}$ can be computed by:
\begin{equation}
\begin{aligned}
A &= \begin{bmatrix}
\Delta R_{kk+1}^T & 0_{3 \times 3} & 0_{3 \times 3} \\
-\Delta R_{ik}(a_k^{\wedge})\Delta t & I_{3 \times 3} & 0_{3 \times 3} \\
-1/2\Delta R_{ik}(a_k^{\wedge})\Delta t^2 & I_{3 \times 3}\Delta t & I_{3 \times 3}
\end{bmatrix},\\
B &= [B_g, B_a], \\
B_g &= \begin{bmatrix}
J_r^k\Delta t \\
0_{3 \times 3} \\
0_{3 \times 3}
\end{bmatrix}, \quad B_a = \begin{bmatrix}
0_{3 \times 3} \\
\Delta R_{ik}\Delta t \\
1/2\Delta R_{ik}\Delta t^2
\end{bmatrix},
\end{aligned}
\end{equation}
$\cdot^{\wedge}$ denotes the skew matrix, $\Sigma \in R^{9\times 9}$ is the covariance matrix of the IMU preintegration term. $\Sigma^{gyro}$ and $\Sigma^{acc}$ are the measurement covariance, which is obtained with the uncertainty prediction of the network. $J_r^k$ is the right Jacobian of $SO(3)$ evaluated with the integrated rotation at the $k$-th time step.

\noindent\textbf{Training Losses} Given the IMU preintegratiown terms $\bm{\gamma}^{b_i}_{b_j}, \bm{\beta}^{b_i}_{b_j}, \bm{\alpha}^{b_i}_{b_j}$ from frame $b_i$ to $b_j$, the $j$-th state of the robot in the world frame $\{W\}$ can be computed by:
\begin{equation}
\label{formula:world_inte}
\begin{aligned}
\mathbf{q}^{w}_{b_j} &= \mathbf{q}^{w}_{b_i} \bm{\gamma}_{b_j}^{b_i}, \\
\mathbf{v}_{b_j}^{w} &= \mathbf{v}_{b_i}^{w} + \mathbf{g}^{w} \Delta t_{ij} + \mathbf{R}^{w}_{b_i} \bm{\beta}_{b_j}^{b_i}, \\
\mathbf{p}^{w}_{b_{j}} &= \mathbf{p}^{w}_{b_i} + \mathbf{v}^{w}_{b_i} \Delta t_{ij} + \frac{1}{2} \mathbf{g}^{w} \Delta t_{ij}^2 + \mathbf{R}^{w}_{b_i} \bm{\alpha}_{b_j}^{b_i},
\end{aligned}
\end{equation}

The correction loss is then designed as:
\begin{equation}
\begin{aligned}
L_r &= \| \text{Log}(\hat{\mathbf{q}}_w^{b_i} \mathbf{q}^w_{b_i}) \|_h, \\
L_v  &= \| \mathbf{v}^{w}_{bi}-\hat{\mathbf{v}}^{w}_{bi} \|_h, \\
L_p  &= \| \mathbf{p}^{w}_{bi}-\hat{\mathbf{p}}^{w}_{bi} \|_h,
\end{aligned}
\end{equation}
where $\hat{\mathbf{q}}_w^{b_i}$, $\hat{\mathbf{v}}^{w}_{bi}$, and $\hat{\mathbf{p}}^{w}_{bi}$ are ground truth. The $\|\cdot\|_h$ denotes the Huber function.

The covariance loss $L^{\text{cov}}$ adopts the Gaussian negative-log-likelihood form defined in main paper Eq.~\maineqref{formula:cov_loss}. Its residual covariances $\Sigma^r_{i,j}$, $\Sigma^v_{i,j}$, $\Sigma^p_{i,j}$ are computed from the IMU preintegration covariances $\Sigma^{\Delta}_{i,j}$ as:
\begin{equation}
\begin{aligned}
\Sigma^r_{i,j} &= J_r^{-1}(\text{Log}(\hat{\mathbf{q}}_w^{b_j} \mathbf{q}^w_{b_j})) \Sigma^{\Delta R}_{i,j} J_r^{-1}(\text{Log}(\hat{\mathbf{q}}_w^{b_j} \mathbf{q}^w_{b_j}))^T, \\
\Sigma^v_{i,j} &= \mathbf{R}^w_{b_i} \Sigma^{\Delta v}_{i,j} (\mathbf{R}^w_{b_i})^T, \\
\Sigma^p_{i,j} &= \mathbf{R}^w_{b_i} \Sigma^{\Delta p}_{i,j} (\mathbf{R}^w_{b_i})^T.
\end{aligned}
\end{equation}

\section{Pose Estimation Improvement via VI Fusion}
\label{app:pose_comparison}

\begin{table}[!t]
    \caption{Pose estimation comparison on EuRoC between MAC-VO (vision only) and MAC-I$^2$ (visual-inertial fusion).}
    \label{tab:app_pose_compare}
    \centering
    \resizebox{\linewidth}{!}{
    \scriptsize
    \begin{tabular}{l l cccccc}
        \toprule
        \textbf{Metrics} & {Methods} & {MH02} & {MH04} & {V101} & {V103} & {V202} & {Avg} \\
        \midrule
        \multirow{3}{*}{\shortstack{\textbf{Rotation}\\\textbf{Error ($\degree$)}}}
        & MAC-VO
        & 0.211 & 0.189 & 0.195 & 0.541 & 0.459 & 0.319 \\
        & MAC-I$^2$ w/o Learned IMU
        & \underline{0.132} & \underline{0.145} & \underline{0.185} & \underline{0.410} & \underline{0.383} & \underline{0.251} \\
        & MAC-I$^2$ w/ Learned IMU
        & \textbf{0.082} & \textbf{0.095} & \textbf{0.159} & \textbf{0.317} & \textbf{0.313} & \textbf{0.193} \\
        \midrule
        \multirow{3}{*}{\shortstack{\textbf{Translation}\\\textbf{Error (m)}}}
        & MAC-VO
        & 0.010 & 0.033 & \textbf{0.008} & 0.043 & 0.024 & 0.024 \\
        & MAC-I$^2$ w/o Learned IMU
        & \textbf{0.009} & \textbf{0.031} & \textbf{0.008} & \underline{0.038} & \underline{0.021} & \underline{0.021} \\
        & MAC-I$^2$ w/ Learned IMU
        & \textbf{0.009} & \textbf{0.031} & \textbf{0.008} & \textbf{0.035} & \textbf{0.019} & \textbf{0.020} \\
        \bottomrule
    \end{tabular}
    }
\end{table}

Table~\ref{tab:app_pose_compare} compares pose estimates produced by our VI initialization against MAC-VO alone on EuRoC. For each initialization segment, the pose error is computed as the average rotation or translation error across all keyframes; the values are then averaged over all segments per sequence.

Our framework consistently improves pose accuracy over MAC-VO across all sequences. The improvement is particularly pronounced in rotation estimation, where MAC-I$^2$ w/ Learned IMU reduces the average rotation error from 0.319$\degree$ to 0.193$\degree$ (39.5\%), as the IMU gyroscope provides accurate short-term rotational information that complements visual estimates. Even without the learned IMU model, our method achieves a 21.3\% reduction in rotation error. Translation improvements are more moderate (0.024~m to 0.020~m, 16.7\%), because MAC-VO already provides strong visual translation constraints and the optimizer naturally assigns more trust to the more accurate source.

\begin{figure}[t]
\centering
\includegraphics[width=\columnwidth]{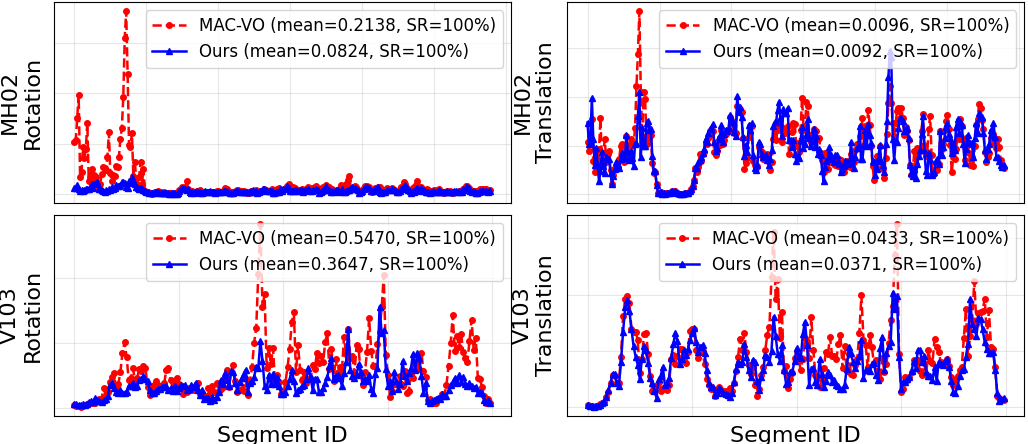}
\caption{Per-segment pose error comparison between MAC-VO and our method on EuRoC MH02 (top) and V103 (bottom). In segments where MAC-VO exhibits large errors, our method produces more accurate estimates by shifting trust toward IMU measurements through covariance weighting.}
\label{fig:app_pose_compare}
\end{figure}

Fig.~\ref{fig:app_pose_compare} visualizes per-segment pose errors on two representative sequences. On MH02, MAC-VO suffers from occasional large rotation errors, while our method consistently maintains low errors. On V103, which involves aggressive motions and challenging visual conditions, MAC-VO exhibits frequent rotation error spikes. In these segments, the visual pose covariance increases to reflect the degraded visual quality, causing the optimizer to assign greater trust to the IMU and thereby producing more robust rotation estimates.

\section{Covariance Generalizability}
\label{app:cov_generalizability}

\subsection{Visual Covariance Generalization}

\begin{figure}[t]
\centering
\subfloat[TartanAirV2 (unseen synthetic scenes, pinhole camera).\label{fig:app_tartanairv2_pose_cov}]{%
    \includegraphics[width=\columnwidth]{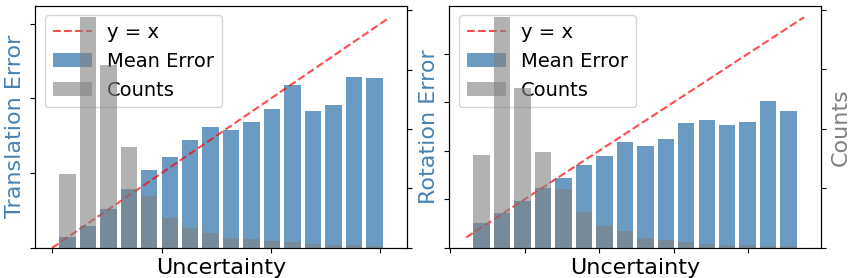}}\\[0.3em]
\subfloat[TUM-VI (real-world, fisheye camera, handheld motion).\label{fig:app_tumvi_pose_cov}]{%
    \includegraphics[width=\columnwidth]{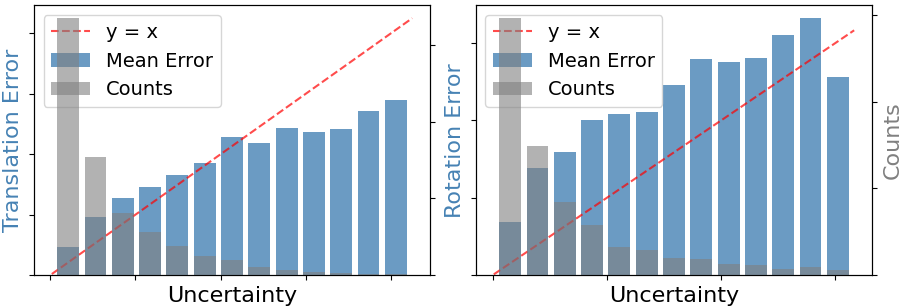}}
\caption{Visual pose covariance binning analysis on (a)~TartanAirV2 and (b)~TUM-VI. Despite being trained only on synthetic TartanAir with pinhole cameras, the covariance model remains metrics-aware across diverse scenes, sensor types, and motion regimes.}
\label{fig:app_vis_cov_generalization}
\end{figure}

Our visual covariance model is trained entirely on synthetic TartanAir with pinhole cameras and applied to real-world datasets \textit{without any fine-tuning}. Fig.~\ref{fig:app_vis_cov_generalization} evaluates generalization on two additional datasets. Fig.~\ref{fig:app_tartanairv2_pose_cov} shows the binning analysis on TartanAirV2, which features diverse unseen synthetic scenes; the predicted uncertainty closely follows the ideal $y = x$ relationship, confirming generalization across scene types. Fig.~\ref{fig:app_tumvi_pose_cov} presents results on TUM-VI, a more challenging domain with fisheye cameras and handheld motions. Despite this domain gap, the covariance model remains metrics-aware and reflects actual error magnitudes.

\subsection{IMU Covariance Generalization}

\begin{figure}[t]
\centering
\includegraphics[width=\columnwidth]{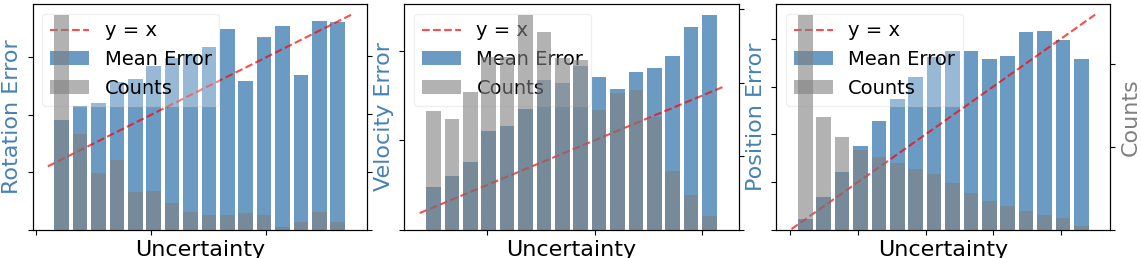}
\caption{IMU covariance binning analysis on held-out TUM-VI room sequences. The predicted covariance remains metrics-aware on a different sensor and motion regime than EuRoC.}
\label{fig:app_tumvi_imu_cov}
\end{figure}

The learned IMU model captures sensor-specific bias and noise characteristics, requiring per-sensor training analogous to standard IMU calibration. Within this setup, the model generalizes across diverse motion regimes on the same sensor. Fig.~\ref{fig:app_tumvi_imu_cov} shows the IMU covariance binning analysis on held-out TUM-VI room sequences, which involve handheld motions distinct from the EuRoC MAV trajectories. The predicted covariance remains well-calibrated and tracks actual integration errors. For scenarios where sensor-specific training data is unavailable, our framework provides a raw-IMU workflow that still achieves strong performance.

\section{Sensitivity to Keyframe Count}
\label{app:sensitivity}

\begin{table}[!t]
    \caption{Sensitivity to keyframe count on EuRoC.}
    \label{tab:app_sensitivity}
    \centering
    \resizebox{\linewidth}{!}{
    \scriptsize
    \begin{tabular}{c c l cccccc}
        \toprule
        \textbf{KFs} & \textbf{Metrics} & {Methods} & {MH02} & {MH04} & {V101} & {V103} & {V202} & {Avg} \\
        \midrule
        \multirow{4}{*}{5}
        & \multirow{2}{*}{\shortstack{Vel\\(m/s)}}
        & w/ Learned IMU
        & 0.012 & 0.029 & 0.011 & 0.031 & 0.025 & 0.022 \\
        & & w/o Learned IMU
        & 0.012 & 0.025 & \textbf{0.010} & 0.029 & 0.022 & 0.020 \\
        \cmidrule{2-9}
        & \multirow{2}{*}{\shortstack{G.Dir\\($\degree$)}}
        & w/ Learned IMU
        & 0.267 & 0.432 & 0.395 & 0.785 & 0.999 & 0.576 \\
        & & w/o Learned IMU
        & 0.842 & 0.854 & 0.914 & 1.197 & 0.871 & 0.936 \\
        \midrule
        \multirow{4}{*}{10}
        & \multirow{2}{*}{\shortstack{Vel\\(m/s)}}
        & w/ Learned IMU
        & \textbf{0.010} & 0.023 & \textbf{0.010} & 0.025 & 0.024 & 0.018 \\
        & & w/o Learned IMU
        & 0.011 & \textbf{0.022} & \textbf{0.010} & 0.025 & \textbf{0.020} & 0.018 \\
        \cmidrule{2-9}
        & \multirow{2}{*}{\shortstack{G.Dir\\($\degree$)}}
        & w/ Learned IMU
        & 0.166 & 0.208 & \textbf{0.332} & 0.581 & 0.804 & 0.418 \\
        & & w/o Learned IMU
        & 0.820 & 0.800 & 0.872 & 1.149 & 0.804 & 0.889 \\
        \midrule
        \multirow{4}{*}{20}
        & \multirow{2}{*}{\shortstack{Vel\\(m/s)}}
        & w/ Learned IMU
        & \textbf{0.010} & 0.023 & 0.011 & 0.024 & 0.024 & 0.018 \\
        & & w/o Learned IMU
        & 0.011 & \textbf{0.022} & \textbf{0.010} & \textbf{0.023} & \textbf{0.020} & \textbf{0.017} \\
        \cmidrule{2-9}
        & \multirow{2}{*}{\shortstack{G.Dir\\($\degree$)}}
        & w/ Learned IMU
        & \textbf{0.138} & \textbf{0.183} & 0.348 & \textbf{0.556} & \textbf{0.775} & \textbf{0.400} \\
        & & w/o Learned IMU
        & 0.827 & 0.802 & 0.816 & 1.083 & 0.821 & 0.870 \\
        \bottomrule
    \end{tabular}
    }
\end{table}

We evaluate sensitivity to the number of keyframes by testing with 5, 10, and 20 keyframes on EuRoC (Table~\ref{tab:app_sensitivity}). All configurations achieve 100\% success rate regardless of keyframe count. Performance improves with more keyframes but with diminishing returns: for gravity direction error with the learned IMU, increasing from 5 to 10 keyframes yields a 27.4\% reduction (0.576$\degree$ $\to$ 0.418$\degree$), while further increasing to 20 brings only a 4.3\% reduction (0.418$\degree$ $\to$ 0.400$\degree$). A similar trend is observed for velocity RMSE and for the workflow without the learned IMU model.

\ifdefined\rootloaded\else\end{document}\fi

\end{document}